\pdfoutput=1
\documentclass[11pt]{article}

\usepackage[final]{acl}
\usepackage{times}
\usepackage{authblk} 

\usepackage{latexsym}
\usepackage{xcolor}
\usepackage{pifont}
\usepackage[T1]{fontenc}
\usepackage[utf8]{inputenc}

\usepackage{microtype}

\usepackage{enumitem}

\usepackage{inconsolata}
\usepackage{microtype}
\usepackage{graphicx}
\usepackage{subfigure, subcaption}
\usepackage{booktabs} % for professional tables
\usepackage{caption}
\usepackage{bm}
\usepackage{hyperref}
\usepackage{makecell}

\usepackage{amsmath}
\usepackage{amssymb}
\usepackage{mathtools}
\usepackage{amsthm}
\usepackage{caption,subfigure,graphicx,epstopdf,multirow,multicol,booktabs,verbatim,wrapfig,bm}

\usepackage[capitalize,noabbrev]{cleveref}

\theoremstyle{plain}

\theoremstyle{definition}
\newtheorem{definition}{Definition}

\theoremstyle{remark}

\usepackage{titletoc}

\usepackage[textsize=tiny]{todonotes}

\title{Towards Tracing Trustworthiness Dynamics: Revisiting Pre-training Period of Large Language Models}

\author{%
\textbf{Chen Qian}\textsuperscript{1,2{$\star$}},
\textbf{Jie Zhang}\textsuperscript{1,3{$\star$}}, 
\textbf{Wei Yao}\textsuperscript{1,2{$\star$}}, 
\textbf{Dongrui Liu}\textsuperscript{1,4}, 
\\ % Line break
\textbf{Zhenfei Yin}\textsuperscript{1,5}, 
\textbf{Yu Qiao}\textsuperscript{1}, 
\textbf{Yong Liu}\textsuperscript{2}$^{\dag}$, 
\textbf{Jing Shao}\textsuperscript{1}$^{\dag}$\\
$^1$ Shanghai Artificial Intelligence Laboratory \\
$^2$ Renmin University of China~
$^3$ University of Chinese Academy of Sciences \\
$^4$ Shanghai Jiao Tong University~
$^5$ The University of Sydney \\

\tt\footnotesize\{qianchen2022, wei.yao, liuyonggsai\}@ruc.edu.cn~~zhangjie@iie.ac.cn~~shaojing@pjlab.org.cn\\
}

\begin{document}
\maketitle
% \begin{NoHyper}
% \def\thefootnote{$\star$}\footnotetext{Equal contribution}
% \def\thefootnote{\dag}\footnotetext{Corresponding author}
% \def\thefootnote{\arabic{footnote}}
% \end{NoHyper}
\let\thefootnote\relax\footnotetext{$^\star$ Equal contribution\hspace{3pt} \hspace{5pt}$^{\dag}$ Corresponding author\hspace{5pt}}

\begin{abstract}

Ensuring the trustworthiness of large language models (LLMs) is crucial.
Most studies concentrate on fully pre-trained LLMs to better understand and improve LLMs' trustworthiness.
In this paper, to reveal the untapped potential of pre-training, we pioneer the exploration of LLMs' trustworthiness during this period, focusing on five key dimensions: reliability, privacy, toxicity, fairness, and robustness. 
To begin with, we apply linear probing to LLMs.
The high probing accuracy suggests that \textit{LLMs in early pre-training can already distinguish concepts in each trustworthiness dimension}.
Therefore, to further uncover the hidden possibilities of pre-training, we extract steering vectors from a LLM's pre-training checkpoints to enhance the LLM's trustworthiness.
Finally, inspired by~\citet{choi2023understanding} that mutual information estimation is bounded by linear probing accuracy, 
we also probe LLMs with mutual information to investigate the dynamics of trustworthiness during pre-training. 
We are the first to observe a similar two-phase phenomenon: fitting and compression~\citep{shwartz2017opening}.
This research provides an initial exploration of trustworthiness modeling during LLM pre-training, seeking to unveil new insights and spur further developments in the field.
Our code is publicly accessible at \url{https://github.com/ChnQ/TracingLLM}.
\end{abstract}

\section{Introduction}

As the capabilities of LLMs increase, their trustworthiness becomes a focal point of widespread attention. 
Guided by global AI governance~\cite{AI_Act,NIST,CLTC} and trustworthy AI~\cite{doi/10.2759/346720,Liu2023trustworthy_ai}, trustworthy LLMs have developed some common categories, especially focusing on five dimensions: reliability, toxicity, privacy, fairness, and robustness~\cite{wang2024decodingtrust,sun2024trustllm}. Delving into LLMs across all these trustworthiness dimensions is essential for society.

To seek a deeper exploration of language models, one of the prominent methods is probing~\citep{zhao2023explainability,rauker2023toward}, which involves training a classifier on the model's representations to identify linguistic and semantic properties acquired by the model~\citep{tenney2019you,pimentel2020information,li2021implicit,belinkov2022probing,rauker2023toward,gurnee2023language,slobodkin2023curious}. In particular, considering trustworthiness, recent attempts reveal that LLM representations contain linearly separable patterns~\citep{zou2023representation,li2023inferencetime,azaria2023internal}.
Unfortunately, existing research has largely focused on fully pre-trained LLMs~\cite{touvron2023llama1}, including those aligned~\cite{ouyang2022training} through Supervised Fine-Tuning (SFT) or Reinforcement Learning from Human Feedback (RLHF). 
This perspective neglects the pre-training period in the context of LLM trustworthiness. 
To our best knowledge, two aspects still remain mysterious: 1) how LLMs dynamically encode trustworthiness during pre-training, and 2) how to harness the pre-training period for more trustworthy LLMs.

%% fig: all ckpts
\begin{figure*}[t]
  \centering
  \subfigure{
    \includegraphics[width=\linewidth]{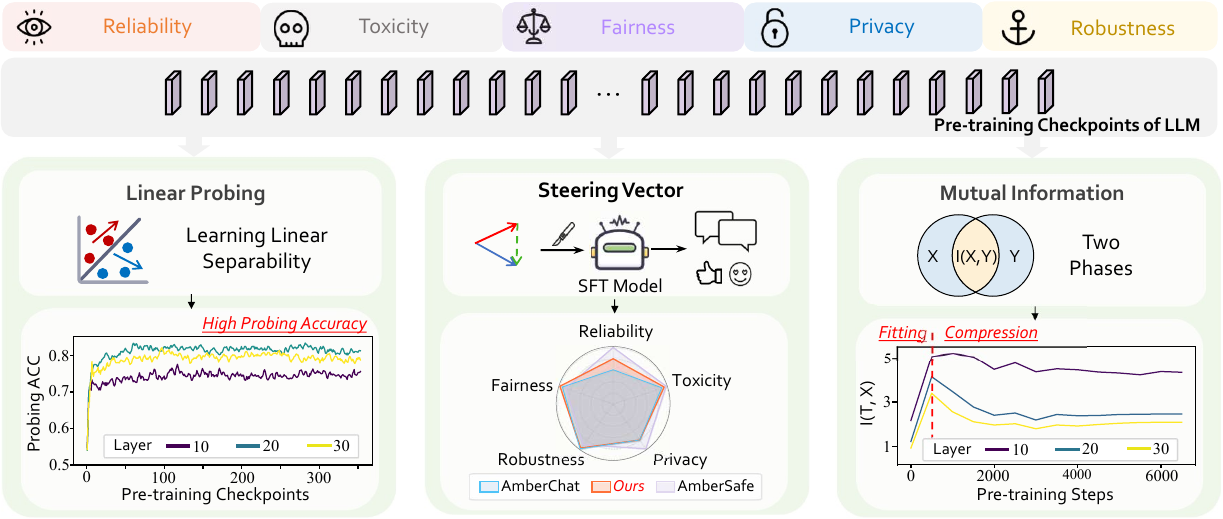}
    \vspace{-15pt}
    \label{fig:sub1}
    }
  \caption{Overview of tracing trustworthiness dynamics during pre-training. 1) Linear probing identifies linearly separable opposing concepts during early pre-training; 2) Steering vectors are developed to enhance LLMs' trustworthiness; 3) Probing LLMs with mutual information reveals a two-phase trend regarding trustworthiness.}
  \label{fig:overview}
  \vspace{-10pt}
\end{figure*}

To address the above issues, we start by analyzing the pre-training dynamics about the trustworthiness of LLM.
More specifically, we use linear probing~\citep{alain2016understanding,belinkov2022probing} across the 360 pre-training checkpoints from LLM360~\citep{liu2023llm360} to explore five dimensions of trustworthiness: 
reliability, toxicity, privacy, fairness, and robustness. 
Our probing results suggest that \textit{after the early pre-training period, middle layer representations of LLMs have already developed linearly separable patterns about trustworthiness}. Such patterns are capable of discerning opposing concepts within each trustworthiness dimension (e.g., discriminating true and false statements). 
Building upon the above observations, we raise an intriguing question: 
\textit{can the pre-training period of an LLM be utilized to enhance its trustworthiness after pre-training?}

We provide insightful answers to the above question by exploring the potential of pre-training checkpoints for better trustworthiness. 
Notably, recent advancements have introduced ``activation intervention,'' a novel suite of techniques for directing language models towards enhanced LLMs' performance by adjusting activations during inference~\cite{turner2023activation, li2023inferencetime, rimsky2023steering, wang2023backdoor}. 
Inspired by these works and the observation of linearly separable patterns in trustworthiness concepts during the LLM's pre-training period, we make preliminary attempts to extract steering vectors from LLM's checkpoints during pre-training, employing them to intervene in the SFT model for trustworthiness enhancement. 
Extensive experiments reveal that \textit{these steering vectors extracted from pre-training checkpoints could promisingly enhance the SFT model's trustworthiness}.
More crucially, these steering vectors achieve a trustworthiness performance that matches or promisingly exceeds that of vectors extracted directly from the SFT model itself.
Our findings introduce novel insights into using pre-training checkpoints for LLM alignment, revealing untapped potential and offering a fresh perspective on enhancing LLM trustworthiness.

Finally, motivated by the theoretical result~\citep{choi2023understanding} that mutual information estimation is bounded by linear probing accuracy, we take an alternative view by probing LLMs with mutual information during pre-training.
To our best knowledge, we are the first to notice that \textit{during the pre-training period of LLMs, there exist two distinct phases regarding trustworthiness: fitting and compression}, which is in line with previous research on traditional DNNs~\citep{shwartz2017opening,noshad2019scalable}.

\section{Probing LLM Pre-training Dynamics in Trustworthiness}

In this section, we probe LLMs to analyze the dynamics of pre-training about trustworthiness.
To begin with, we describe the datasets for each trustworthiness dimension in Section~\ref{subsec:trust_dim_data}.
Then, we introduce the experimental setup in Section~\ref{sec:methods}.
The probing results in Section~\ref{subsec:probing_results} suggest that middle-layer LLM representations from early pre-training have already exhibited linearly separable patterns.

\begin{figure*}[ht]
  \centering
\includegraphics[width=\linewidth]{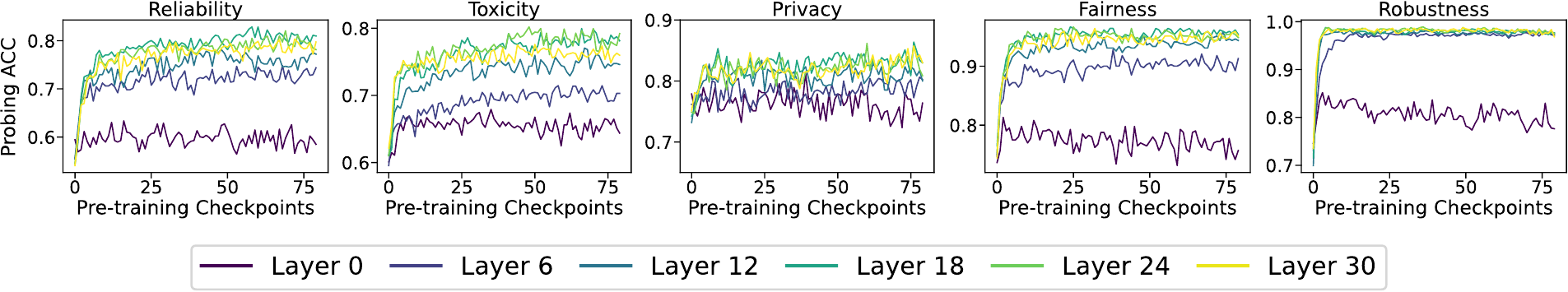}
\vspace{-15pt}
\caption{The linear probe accuracy on five trustworthiness dimensions for the first 80 pre-training checkpoints. For each checkpoint, we report the results from layers \{0, 6, 12, 18, 24, 30\}. The results from all layers of the 360 checkpoints are in Appendix~\ref{appendix:full-probing}.
}
\label{fig:tqa-llm360-probing}
\vspace{-5pt}
\end{figure*}

\subsection{Research Dimensions and Datasets of Truthworthy LLM} \label{subsec:trust_dim_data}

Existing research in AI governance and trustworthy AI provides guidance for establishing comprehensive and reliable dimensions of trustworthy LLMs in this study. 
Governments~\cite{NIST,doi/10.2759/346720}, organizations~\cite{AI_Act,AI_Verify}, and research institutions~\cite{CLTC,liu2023trustworthy} worldwide have proposed classifications from various perspectives such as the AI lifecycle, the acceptability of AI risk, considering AI governance at different levels including individual, institutional, and societal. 
Among these, categories stemming from the technological aspect offer guidance for trustworthy AI~\cite{Liu2023trustworthy_ai}, such as robustness, fairness, accountability, transparency, etc.
% Existing research on AI governance~\cite{NIST,doi/10.2759/346720,AI_Act} and trustworthy AI~\cite{Liu2023trustworthy_ai,AI_Verify} lays the groundwork for developing a comprehensive understanding of trustworthy LLMs. 
Guided by these principles, various studies classify trustworthy LLMs from different perspectives, yet some dimensions consistently emerge across these works~\cite{liu2023trustworthy,wang2024decodingtrust,sun2024trustllm}. Therefore, we delve into five of these key dimensions: reliability, toxicity, privacy, fairness, and robustness, employing canonical datasets for each to support our study.

% \noindent$\bullet$
\noindent
\textbf{Reliability.}
TruthfulQA~\cite{Lin_Hilton_Evans_2022}, a benchmark dataset for evaluating LLMs' truthfulness discernment~\cite{touvron2023llama}, includes 817 questions across 38 categories aimed at assessing the veracity of model-generated answers.

% \noindent$\bullet$
\noindent
\textbf{Toxicity.}
ToxiGen~\cite{hartvigsen2022toxigen} is a broad dataset featuring implicit toxic and non-toxic statements across 13 minority demographics, enabling toxicity modeling assessment in LLMs.

% \noindent$\bullet$
\noindent
\textbf{Privacy.}
We choose the tier 2 tasks from ConfAIde~\cite{mireshghallah2023llms} to assess LLMs' privacy awareness, with ConfAIde targeting contextual privacy and identifying vulnerabilities in LLMs' privacy reasoning.

% \noindent$\bullet$
\noindent
\textbf{Fairness.}
We use StereoSet~\citep{stereoset} to measure the stereotype modeling ability, i.e., whether LLMs capture stereotypical biases about race, religion, profession, and gender.

% \noindent$\bullet$
\noindent
\textbf{Robustness.}
We introduce typos by randomly changing the case of 5\% letters in each sentence from SST-2~\citep{socher2013recursive} from GLUE benchmark~\citep{wang2018glue}. The original sentence, as well as the corresponding perturbed sentence, are synthesized into a new dataset.

For each dataset above, we assign a label to every sentence based on whether it is trustworthy, i.e., truthful, toxic, privacy-aware, fair, and perturbed. 
We maintain a balanced dataset for each trustworthiness dimension. 
Further details are available in Appendix~\ref{appendix:dataset}.

\subsection{Experimental Setup} \label{sec:methods}
\paragraph{The models under study.}
We investigate the pre-training period of LLMs through the 360 pre-training checkpoints provided by LLM360~\citep{liu2023llm360}. 
Simultaneously, they also release an instruction fine-tuned conversational model named AmberChat and an aligned conversational model named AmberSafe. The models mentioned are all of the 7B parameter scale.

\paragraph{Activation dataset.}
Given each original dataset consisting of sentences and the corresponding class labels, we feed the sentence into LLMs and collect the corresponding activations of the last token~\cite{li2023inferencetime,gurnee2023language} for each layer. The activation dataset $\mathcal{D}=\{( \mathbf{x}_i ,y_i) \}_{i=1}^N$ is constructed with 
% contextualized token representations
the activations $\mathbf{x}_i \in R^d$ and the corresponding binary labels $y_i \in \{0,1\}$.

\paragraph{Linear probing.}
We employ the linear probing method~\citep{alain2016understanding,tenney2019you,pimentel2020information,li2021implicit,belinkov2022probing} to analyze the activation datasets. For each trustworthiness dataset, every layer of each pre-training checkpoint within LLM360 produces an activation dataset. Therefore, there are $360 \times 32$ activation datasets for all 32 layers across 360 checkpoints. We randomly split each activation dataset into training and test sets by 4:1, and fit a binary linear classifier on the training set. 
We train a classifier for each activation dataset, which yields $360 \times 32$ classifiers. We report the accuracy on the test set.

\subsection{Probing Results} \label{subsec:probing_results}

\paragraph{Middle layer representations exhibit linearly separable patterns.}
For each checkpoint during pre-training, Figure~\ref{fig:tqa-llm360-probing} shows that the accuracy is relatively higher for middle layers (the 12-th and 18-th layers). The full results in Appendix~\ref{appendix:full-probing} also support such characteristic of middle layers (about the 18-th layer).
It inspires us that the representations from middle layers exhibit rich linear encoded information to distinguish those different concepts.
Also, the observation meets with other literature considering linear probing in the era of LLMs~\citep{li2023inferencetime, zou2023representation, burns2022discovering}, which also empirically validates the capability of middle layers. 
Moreover, a similar phenomenon has also been found in earlier linear probing literature for BERT~\citep{hewitt2019structural, van2019does}, which may implicitly suggest some similarity between LLMs and relatively small pre-trained models.

\paragraph{The potential of pre-training checkpoints.}
Figure \ref{fig:tqa-llm360-probing} shows that for each layer over the whole pre-training period, the probing accuracy increases during the initial pre-training phase, followed by fluctuation throughout the remaining pre-training period.
The trend enlightens us that models during the early stages of pre-training can already encode these different concepts well in a simple linear manner. Such trustworthiness concepts are linearly represented in the latent space of LLMs, which supports linear representation hypothesis~\citep{park2023linear} and other empirical study~\citep{zou2023representation}.

%%% fig-pipeline
\begin{figure}[t]
    \centering
    \includegraphics[width=\linewidth]{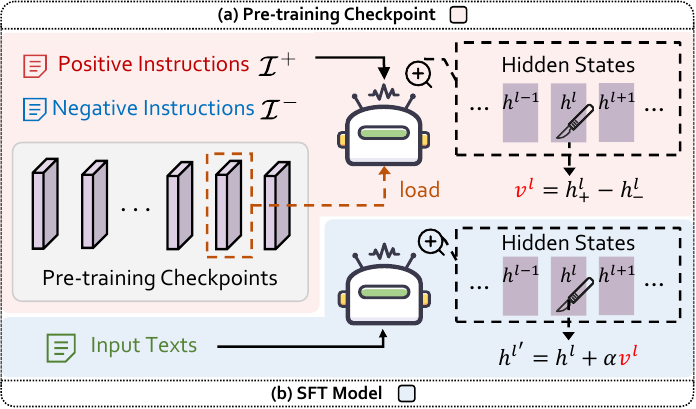}
    \caption{A schematic illustration of 
    (a) constructing a steering vector from the pre-training checkpoints and (b) intervening in the SFT model towards more trustworthiness by employing the steering vector.
    }
    \label{fig:pipeline}
    \vspace{-10pt}
\end{figure}

\section{Controlling Trustworthiness via the Steering Vectors from Pre-training Checkpoints}
\label{sec:steering}
% Having observed impressive linear separability in trustworthiness concepts during the model's pre-training period, 
In this section, we aim to unravel the potential of checkpoints from the pre-training period to assist in enhancing the trustworthiness performance of the SFT model (i.e., AmberChat), based on activation intervention techniques~\cite{turner2023activation, li2023inferencetime,rimsky2023steering}.
We first outline the method of activation intervention on the SFT model using the steering vectors extracted from pre-training checkpoints in Section~\ref{subsec:steering_method}. Next, we introduce the experimental setup in Section~\ref{subsec:steering_setup}. We then explore how steering vectors extracted from pre-training checkpoints enhance performance across distinct dimensions of trustworthiness in Section~\ref{subsec:single-trustworthy}, presenting a series of findings and observations. Finally, we examine using the same techniques to boost the overall trustworthiness performance of the SFT model in Section~\ref{subsec:uni-trustworthy}.

% steering vector - TQA table
\begin{table*}[t!]
\centering
\vspace{-5pt}
\caption{
Results of activation intervention on TruthfulQA, general ability benchmarks, and the other trustworthiness benchmarks.
% compared to baselines.
% using the steering vector extracted from pre-training checkpoints.
The best results are highlighted in \textbf{bold}, and the runner-ups are \underline{underlined}. $\bm v_{ckpt\_{179}}$ and $\bm v_{AmberChat}$ represent AmberChat intervened by steering vectors derived from ckpt\_179 and AmberChat, respectively.}
\vspace{1mm}
\label{table:tqa}
\setlength{\tabcolsep}{1.5mm}
\scalebox{0.70}{
    \begin{tabular}{ll|ccc|cccc|cccc}
    \toprule
    & \multirow{2}{*}{Method} & \multicolumn{3}{c|}{\makecell{TruthfulQA Metrics}} & \multicolumn{4}{c|}{General Abilities} & \multicolumn{4}{c}{Trustworthiness Abilities} \\
    & & Truth$\uparrow$ & Info$\uparrow$ & Truth * Info$\uparrow$ & ARC$\uparrow$ & MMLU$\uparrow$ & MathQA$\uparrow$ & RACE$\uparrow$ & ToxiGen$\downarrow$ & ConfAIde$\uparrow$ & StereoSet$\uparrow$ & SST-2$\uparrow$ \\ 
    \toprule
    \multirow{1}{*}{Baseline}
    % ------------- Baselines rows here -------------
    & AmberChat & 0.3931 & \underline{0.9484} & 0.3728 & \textbf{0.6006} & \textbf{0.3659} & \underline{0.2593} & \underline{0.3904} & 0.0920 & 0.5055 & \textbf{0.5379} & \textbf{0.5757}\\
    % ------------- Fine-tuned rows here -------------
    \midrule
    \multirow{2}{*}{Fine-tuned}
    & Full & 0.4229 & \textbf{0.9602} & 0.4060 & 0.4315 & 0.2355 & 0.2499 & 0.3187 & \textbf{0.0020} & 0.5294 & \underline{0.5031} & \textbf{0.5757} \\ \cmidrule(l){2-13}
    & Lora & 0.3221 & 0.9329 & 0.3004 & 0.5758 & 0.3314 & \textbf{0.2620} & 0.3742 & \underline{0.0080} & \textbf{0.6411} & 0.4980 & \underline{0.5734} \\
    % ------------- Steering Vector rows here -------------
    \midrule
    \multirow{2}{*}[-0.5ex]{\parbox{1cm}{Activation\\Intervention}}
    & \parbox{1.8cm}{$\bm v_{ckpt\_{179}}$} & \textbf{0.7322} & 0.9337 & \textbf{0.6837} & \underline{0.5834} & 0.3358 & 0.2422 & 0.3876 & 0.0360 & \underline{0.6181} & 0.5000 & 0.5229\\ \cmidrule(l){2-13}
    & \parbox{1.8cm}{$\bm v_{AmberChat}$} & \underline{0.6978} & \underline{0.9484} & \underline{0.6618} & 0.5829 & \underline{0.3388} & 0.2482 & \textbf{0.3943} & 0.0320 & 0.5192 & 0.4580 & 0.5367 \\
    \bottomrule
    \end{tabular}
}
\end{table*}

\subsection{Activation Intervention}
\label{subsec:steering_method}
% 1. constractive prompts
Initially, we partition the training dataset into two distinct collections based on the labels, $\mathcal{I}^{+}$ and $\mathcal{I}^{-}$, representing positive instructions and negative instructions, respectively.
Following this partition, we collect the activations of LLM w.r.t. these instructions, denoted by $A_c^l(\mathcal{I}^{+})$ and $A_c^l(\mathcal{I}^{-})$, where $A_c^l$ denotes the function that extracts the activations from the $c$-th checkpoint at $l$-th layer.
% 2. take difference
Subsequently, we compute the centroid of the activations from each set and take their difference to obtain  the ``mass mean vector,''~\cite{li2023inferencetime, marks2023geometry} which serves as our steering vector
\begin{equation}
    \bm v_c^l = \overline{A_c^l}(\mathcal{I}^{+})- \overline{A_c^l}(\mathcal{I}^{-}).
\end{equation}
% In contrast to previous works~\cite{li2023inferencetime, turner2023activation, rimsky2023steering} that construct steering vectors from the model itself, our intent is to construct these vectors from the pre-training checkpoints.
% 3. inference intervention
Finally, we employ the steering vector to intervene in the model's activations, as illustrated below
\begin{equation}
\label{eq-intervene}
\bm h^{l'} = \bm h^{l} + \alpha \bm v_c^{l},
\end{equation}
where $\bm h^{l}$ denotes representation at the $l$-th layer of the model, $\bm h^{l'}$ denotes the corresponding representation after the intervention; $\alpha$ is a rescale hyperparameter that indicates the strength of the intervention. 
Figure~\ref{fig:pipeline} illustrates the schematic diagram of the intervention method.
Note that the intervention described by Eq.~\eqref{eq-intervene} occurs at each step during the autoregressive inference.

\subsection{Experimental Setup}
\label{subsec:steering_setup}
\paragraph{Evaluation on Trustworthiness Datasets.}
For TruthfulQA, we fine-tune two GPT-3 models as ``GPT-judge'' and ``GPT-info'' guided by~\cite{Lin_Hilton_Evans_2022}, to predict the truthfulness and informativeness of the generated outputs from LLMs, respectively.
For ToxiGen, we follow~\cite{touvron2023llama}, employing fine-tuned RoBERTa~\cite{hartvigsen2022toxigen} to evaluate the toxicity of contents generated by LLMs, and finally reporting the proportion of generated text classified as toxic.
For ConfAIde, StereoSet, and perturbed SST-2, with the adaptation of converting possible multiple-choice questions into binary classification tasks, we prompt LLMs to generate choices and then evaluate the accuracy.
Please refer to Appendix~\ref{appendix:exp-setting} for more details.

% steering vector - Toxigen table
\begin{figure}[t!]
  \centering
    \includegraphics[width=0.9\linewidth]{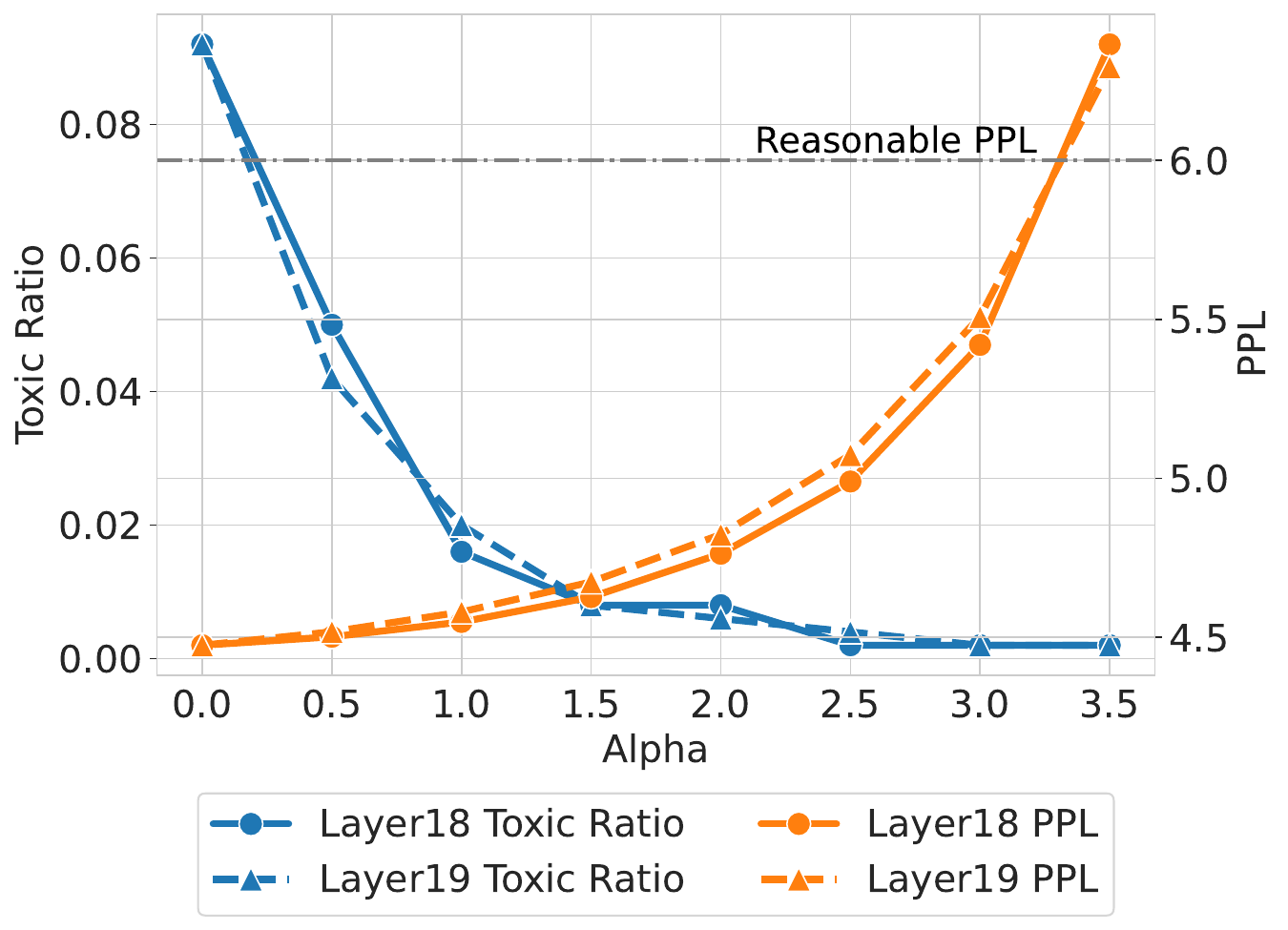}
    \vspace{-5pt}
  \caption{The trends of toxic ratio and PPL as the intervention strength $\alpha$ increases.}
  \label{fig:toxigen_ratio_ppl}
  \vspace{-10pt}
\end{figure}
% \hspace{-10mm}

% \noindent
\paragraph{Details of Steering Vectors Construction.}
For the activation dataset, we consider it from two perspectives: 1) For controlling the performance of individual subcategories under trustworthiness in Section~\ref{subsec:single-trustworthy}, we utilize the corresponding datasets described in Section~\ref{subsec:trust_dim_data}, where the steering vectors are constructed from the development set and no data leakage occurs during the evaluation; 2) For controlling the overall trustworthiness performance in Section~\ref{subsec:uni-trustworthy}, we employ PKU-SafeRLHF-10K\footnote{https://huggingface.co/datasets/PKU-Alignment/PKU-SafeRLHF-10K}, a dataset proposed in~\cite{ji2023beavertails} for RLHF training.
For the checkpoint, we simply select the checkpoint that is halfway through the pre-training process for experiments, namely the checkpoint ckpt\_179, which has already learned linearly separable patterns (i.e., performs a high probing accuracy as shown in Figure~\ref{fig:tqa-llm360-probing}).
Regarding the selection of layer and $\alpha$, we first narrow down the hyperparameter range based on Perplexity (PPL), and then empirically determine the optimal parameters using a coarse-grained grid search~\cite{li2023inferencetime, turner2023activation, wang2023backdoor}.

\subsection{Intervention to Enhance Distinct Trustworthiness Dimensions}
\label{subsec:single-trustworthy}
In this subsection, we present several key observations that illuminate the intricate dynamics of steering vectors in modulating the trustworthiness of the SFT model.

% steering vector - Fairness 
\begin{table*}[ht]
\centering
% \vspace{-3pt}
\caption{Results of activation intervention on StereoSet, general ability benchmarks, and the other trustworthiness benchmarks.
% compared to baselines, using the steering vector extracted from pre-training checkpoints.
Format and significance markers keep consistent with Table~\ref{table:tqa}.
}
\vspace{1mm}
\label{table:fairness}
\setlength{\tabcolsep}{2.5mm}
\scalebox{0.7}{
    \begin{tabular}{ll|c|cccc|cccc}
    \toprule
    & \multirow{2}{*}{Method} & \multicolumn{1}{c|}{\makecell{Fairness  Metric}} & \multicolumn{4}{c|}{General Abilities} & \multicolumn{4}{c}{Trustworthiness Abilities} \\
    & & StereoSet $\uparrow$ & ARC$\uparrow$ & MMLU$\uparrow$ & MathQA$\uparrow$ & RACE$\uparrow$ & TruthfulQA$\uparrow$ & ToxiGen$\downarrow$ & ConfAIde$\uparrow$ & SST-2$\uparrow$ \\ 
    \toprule
    \multirow{1}{*}{Baselines}
    % ------------- Baselines rows here -------------
    & AmberChat & 0.5379 & \textbf{0.6006} & \textbf{0.3659} & \textbf{0.2593} & \textbf{0.3904} & \textbf{0.3728} & 0.0920 & \textbf{0.5055} & \textbf{0.5757} \\
    % ------------- Steering Vector rows here -------------
    \midrule
    \multirow{2}{*}[-0.5ex]{\parbox{1cm}{Activation\\Intervention}}
    & \parbox{1.8cm}{$\bm v_{ckpt\_{179}}$} & \underline{0.5799} & \underline{0.5986} & \underline{0.3524} & 0.2499 & 0.3914 & 0.2851 & \textbf{0.0600} & 0.5055 & 0.5390\\ \cmidrule(l){2-11}
    & \parbox{1.8cm}{$\bm v_{AmberChat}$} & \textbf{0.5830} & 0.5958 & 0.3508 & \underline{0.2519} & \underline{0.3952} & \underline{0.3352} & \underline{0.0820} & 0.5055 & \underline{0.5528} \\
    
    \bottomrule
    \end{tabular}
}
\end{table*}

% fig: person
\begin{figure}
    \centering
    \includegraphics[width=\linewidth]{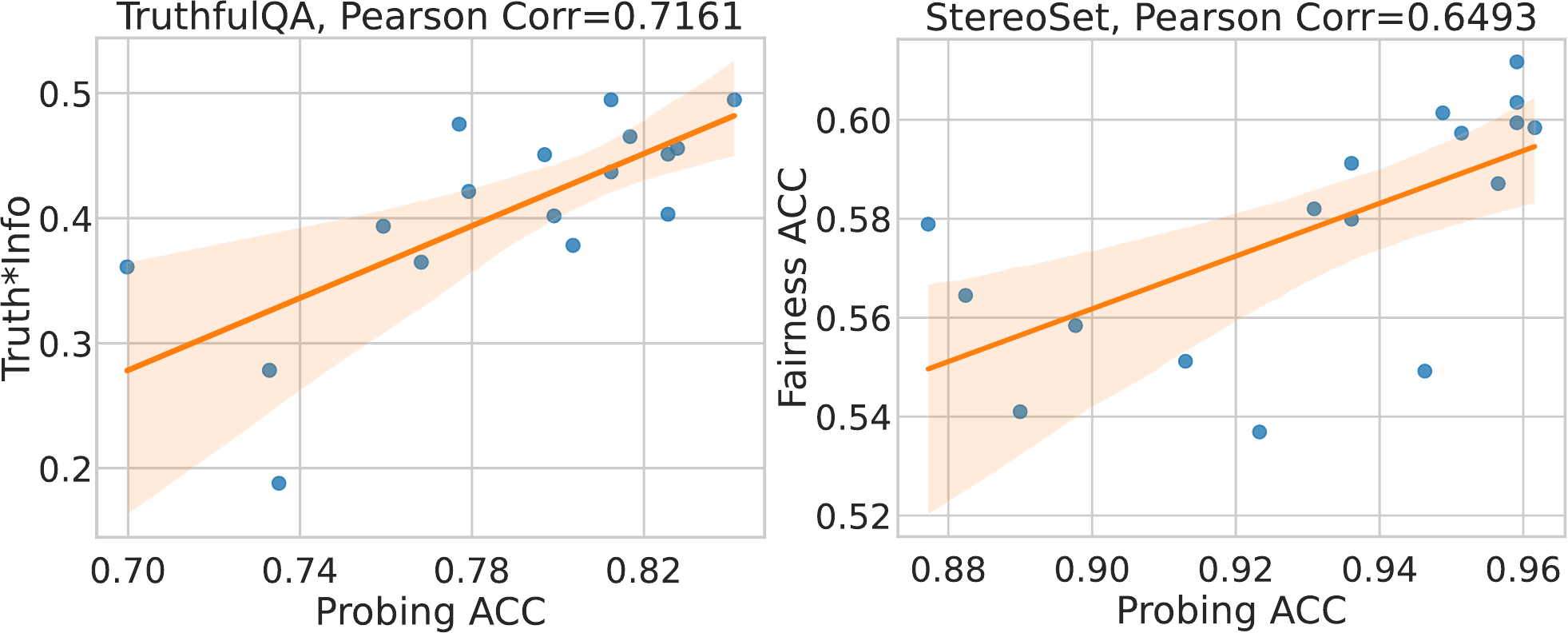}
    \vspace{-10pt}
    \caption{Pearson Correlation Coefficient for Probing ACC and trustworthiness performance.}
    \label{fig:person}
    \vspace{-13pt}
\end{figure}

\noindent
\textbf{Observation 1. }
\textit{Steering vectors derived from pre-training checkpoints could significantly enhance the SFT model's performance in TruthfulQA, ToxiGen, and StereoSet.} For TruthfulQA and StereoSet, clear performance enhancement can be observed in Table~\ref{table:tqa} and Table~\ref{table:fairness}, respectively.  Regarding ToxiGen, when the strength of intervention $\alpha$ is set to 0.5, there is already a reduction of approximately 50\% in the rate of toxic content generation, with a negligible perturbation in perplexity.
Besides, sampling checkpoints from various stages of the pre-training period, we observe a relatively strong linear correlation between the trustworthiness performance and the probing accuracy of pre-training checkpoints in Figure~\ref{fig:person}. 
This suggests that, once the model has developed linearly separable patterns (represents a high probing accuracy) w.r.t. the trustworthiness concepts during the pre-training process, the constructed steering vector may have the potential to positively intervene in the SFT model's trustworthiness.

\noindent
\textbf{Observation 2.}
\textit{Steering vectors derived from pre-training checkpoints and SFT model perform broadly comparable performance yet exhibit variations across various tasks.}
Table~\ref{table:tqa} shows that, compared to the steering vector extracted from AmberChat, the steering vector from the pre-training checkpoint (ckpt\_179) guides the SFT model to exhibit more ``truthfulness.'' Moreover, it performs slightly better on ARC, ConfAIde, and StereoSet, while the opposite is true for other tasks. It is important to note that we only selected a single checkpoint from the pre-training process for experimentation, without undergoing fine-grained hyperparameter selection. Therefore, we believe these pre-training checkpoints hold significant untapped potential for aiding LLMs toward trustworthiness.

% fig: radar
\begin{figure}[t!]
    \centering
    \includegraphics[width=0.95\linewidth]{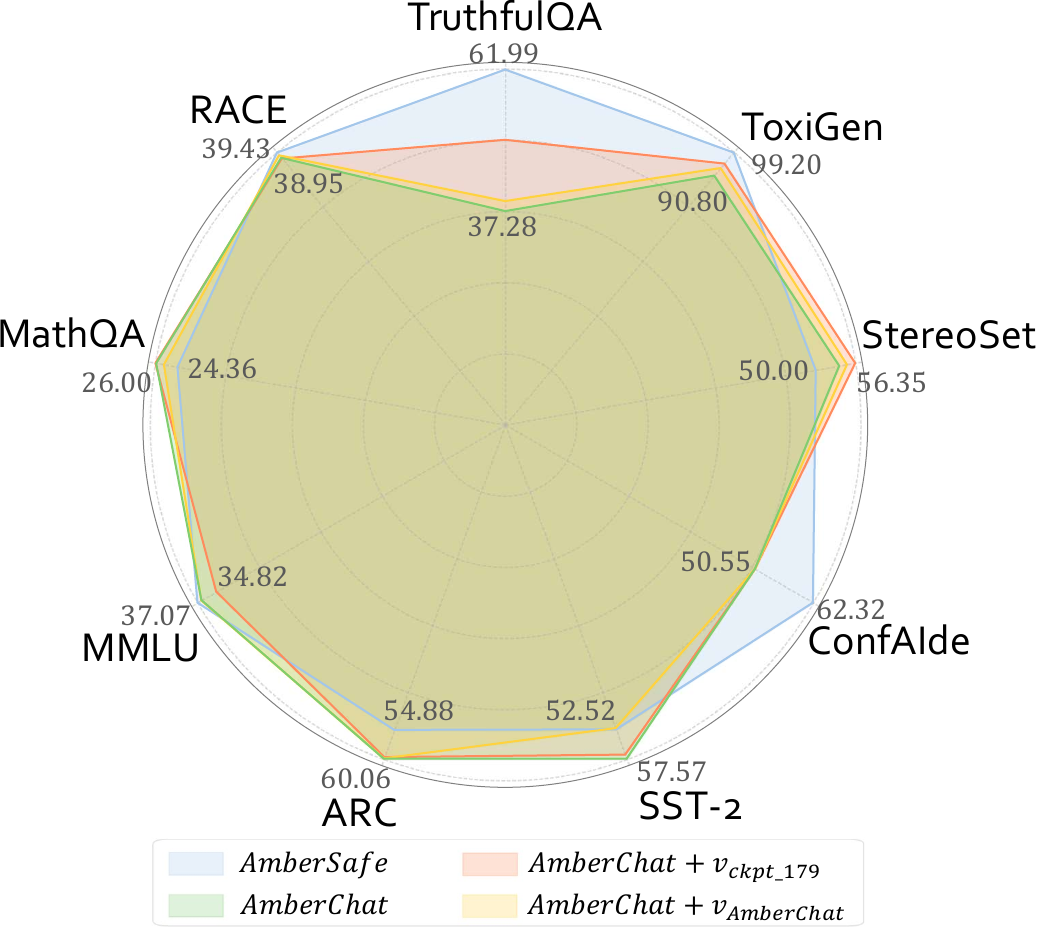}
    \vspace{-5pt}
    \caption[]{Performance of various models across four general capabilities and five trustworthiness capabilities. AmberChat and AmberSafe are fine-tuned models from LLM360. $\bm v_{ckpt\_{179}}$ and $\bm v_{AmberChat}$ represent steering vectors from ckpt\_179 and AmberChat, respectively.}
    \label{fig-radar}
    \vspace{-10pt}
\end{figure}

%%%%%%%%%%%%%%%%%%%%%%%%%%%%%%%%%%%%%%%%%%%
% Observations Here
\noindent
\textbf{Observation 3.}
\textit{Intervening in the model slightly impairs its general capabilities as a marginal cost for trustworthiness enhancement.} We evaluate the model's performance on four common benchmarks for general capabilities, where a trend of slight performance decline is observed after the intervention, as indicated in the ``General Abilities'' part of Tables~\ref{table:tqa} and~\ref{table:fairness}. Additionally, we also observe the impact of the intervention strength $\alpha$ on the generative performance of the model.
Taking ToxiGen as an example, Figure~\ref{fig:toxigen_ratio_ppl} illustrates the relationship between the proportion of toxic content generated by the model and perplexity as the intervention strength $\alpha$ increases. If we continuously increase the intervention strength, although the proportion of toxicity may continue to decline, the perplexity of the model correspondingly increases, manifesting as a tendency to produce meaningless repetitive content or gibberish.

\noindent
\textbf{Observation 4.}
\textit{When the quantity and quality of fine-tuning data are limited, activation intervention by steering vectors may be a more effective approach for the current task. }
We fine-tune the SFT model with positive QA pairs from the training set using both full-parameter fine-tuning and LoRA fine-tuning as a comparison, given that data in TruthfulQA naturally exists in the form of QA pairs. As shown in Table~\ref{table:tqa}, the model fine-tuned with all parameters exhibits only minor improvements on TruthfulQA while experiencing a significant decline in general capabilities. Meanwhile, the fine-tuned model by LoRA demonstrates a noticeable decrease in TruthfulQA, though somewhat preserving performance in general capabilities.

\noindent
\textbf{Observation 5.}
\textit{Trade-offs exist between different dimensions of trustworthiness.} 
% some examples
For instance, as seen in Table~\ref{table:tqa}, while steering vector intervention enhances the model's truthfulness performance, it also compromises performance on fairness and robustness. 
% some previous works
Previous research has witnessed a trade-off between trustworthiness dimensions. 
For example, privacy-fairness trade-off~\citep{privacy_fairness_1}, robustness-privacy trade-off~\citep{robustness_privacy_1}, and robustness-fairness trade-off~\citep{robustness_fairness_1}. 
% For example, privacy-fairness trade-off~\citep{privacy_fairness_1,privacy_fairness_2}, robustness-privacy trade-off~\citep{robustness_privacy_1,robustness_privacy_2}, robustness-fairness trade-off~\citep{robustness_fairness_1,robustness_fairness_2,robustness_fairness_3,robustness_fairness_4}, and the trade-off among fairness definitions~\citep{fair_impossible_1,fair_impossible_2,fair_impossible_3}. 
Similar to \citep{liang2022holistic}, we also suggest that the connection between different trustworthiness dimensions relies on their definitions. Many pairs of trustworthiness in LLMs remain unstudied, and we advocate for future research in this area.
% While prior works may suggest certain relationships in the context of LLMs, many pairs remain unstudied, and we lack definitive evidence of their reliability. 
% We encourage researchers in the future to explore the inter-metric relationships of LLMs regarding trustworthiness.

\subsection{Intervention to Enhance Universal Trustworthiness}
\label{subsec:uni-trustworthy}

In this subsection, we aim to leverage steering vectors to comprehensively enhance the model's trustworthiness.
Unlike Section~\ref{subsec:single-trustworthy} where steering vectors are constructed using datasets from different dimensions of trustworthiness, here we employ a general dataset for alignment (described in Section~\ref{subsec:steering_setup}), which may encompass data across multiple dimensions of trustworthiness.

\noindent
% \paragraph{Steering vectors constructed from pre-training checkpoints by universal alignment datasets could intervene in the model to simultaneously enhance multiple trustworthiness capabilities.}
\paragraph{Trustworthiness enhancement with steering vectors from universal alignment datasets.}
% 1. 描述实验现象
Figure~\ref{fig-radar} suggests that intervening in the SFT model with steering vectors can influence its trustworthiness, showing notable improvements in certain dimensions (which may potentially linked to the characteristics of the datasets employed), with only marginal losses (in ARC, MMLU) or even marginal gains (in MathQA, RACE) in general capabilities. Moreover, steering vectors derived from checkpoints during the pre-training period demonstrate superior effectiveness in enhancing trustworthiness.
For AmberSafe, which employs a substantial cost for alignment, we note its overall best performance (as seen in the blue line), particularly holding a significant advantage in privacy and TruthfulQA. However, it's noteworthy that merely using 10k alignment data to construct steering vectors from a pre-training checkpoint for intervening in the SFT model brings about impressive improvements across various dimensions of trustworthiness, which reveals the untapped potential of pre-training checkpoints in aiding the model towards better trustworthiness.

\section{Probing LLMs using Mutual Information}
Recently,
\citet{choi2023understanding} shows that mutual information estimation is bounded by linear probing accuracy.
Also, the mutual information can be used to investigate the dynamics of neural networks during training~\citep{shwartz2017opening,saxe2019information,goldfeld2020information, pimentel2020information, geiger2021information,lorenzen2021information,zhou2023mcgra}. 
Therefore, motivated by the above, we adopt a different perspective by probing LLM checkpoints through the lens of mutual information, particularly focusing on the aforementioned trustworthiness dimensions. 

We explain our probing strategy and experimental setup in Section~\ref{subsec:information_llm} and Section~\ref{subsec:information_exp}, respectively.
The empirical observations are shown and analyzed in Section~\ref{subsec:information_result}. 
In particular, we find that there is a phase transition from ``fitting'' to ``compression'' during the pre-training period of LLMs, which is consistent with previous studies on traditional DNNs~\citep{shwartz2017opening,noshad2019scalable}.

\subsection{Probing Strategy} \label{subsec:information_llm}
The mutual information between two continuous random variables, $X$ and $Y$, is defined as 
$$I(X , Y)=\int_Y \int_X p(x, y) \log \frac{p(x, y)}{p(x) p(y)} d x d y.$$
It is a measure of the independence between two variables. 
Given the dataset of trustworthiness in Section~\ref{subsec:trust_dim_data}, we represent each dataset using the first layer activation $X$, and $Y$ denotes the corresponding label vector. Additionally, $T$ represents the feature matrix from the target layer of an LLM. Thus, we probe LLMs with $I(T,X)$ and $I(T,Y)$ during pre-training.

In principle, our strategy differs from~\citet{shwartz2017opening} in three ways.
Firstly, we do not use the pre-training dataset of LLMs. 
Instead, we carefully design activation datasets to represent specified trustworthiness properties. Secondly, 
we use the first layer representation to indicate the original dataset because they contains more information than representations from other layers
~\citep{cover1999elements,tishby2015deep,shwartz2017opening}. Finally, we follow~\citet{ma2020hsic} to use HSIC~\citep{gretton2005measuring} as an estimator of mutual information because it is challenging to accurately compute in high dimensions~\citep{kraskov2004estimating,alemi2016deep,poole2019variational}.

%% fig: 信息瓶颈
\begin{figure}[t]
    \centering
    \includegraphics[width=\linewidth]{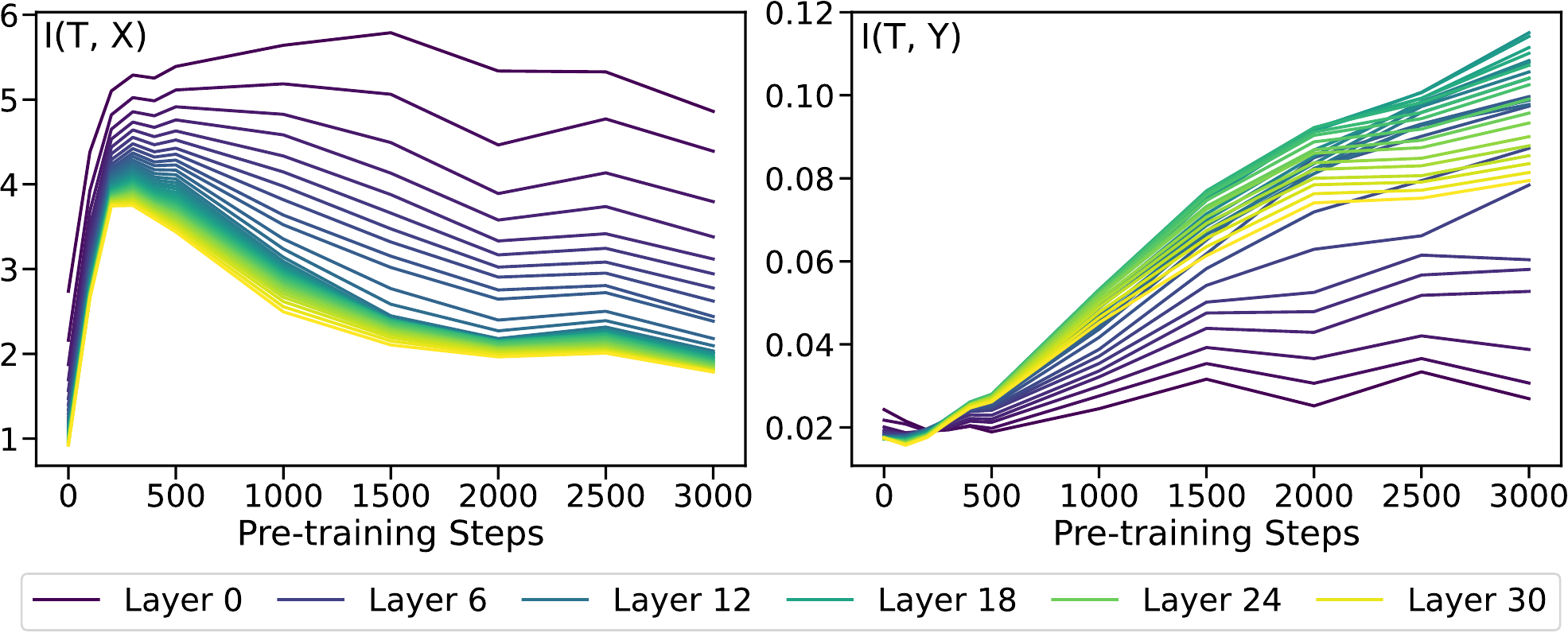}
    \caption{The dynamics of $I(T, X)$ and $I(T, Y)$ for TruthfulQA across various layers during pre-training. A similar trend in other datasets is in Appendix~\ref{appendix:mi_others}.}
    \label{fig-mi-tqa}
    \vspace{-10pt}
\end{figure}

\subsection{Experimental Setup} \label{subsec:information_exp}

Following the official code and reported hyperparameters from \citet{liu2023llm360}, we initiate pre-training from a randomly initialized model using the corpus for the first checkpoint, and save more granular checkpoints to observe finer phenomena. 
More discussions are available in Appendix~\ref{appendix:exp-setting}.

\subsection{The Dynamics of Pre-training} \label{subsec:information_result}

\paragraph{The trend of mutual information.}
Figure~\ref{fig-mi-tqa} shows that $I(T, X)$ generally exhibits an initial increase followed by a decrease across all the considered layers 
during pre-training. 
And $I(T, Y)$ continues to show a consistent upward trend. Note that middle layer representations exhibit a larger $I(T, Y)$ compared to that from other layers. It suggests that middle-layer representations encode more information about the opposing concepts of trustworthiness.

\paragraph{From ``fitting'' to ``compression.''}
Overall, considering $I(T,X)$ and $I(T,Y)$ collectively, it becomes evident that there are two phases during pre-training. In the first and shorter phase, both $I(T, X)$ and $I(T, Y)$ increase. While in the second and much longer phase, $I(T, X)$ decreases and $I(T, Y)$ continues to increase. 
Although our strategy is completely different from~\citet{shwartz2017opening}, the two-phase phenomenon exhibits similarities. 
At the beginning of pre-training, the randomly initialized LLM fails to preserve the relevant information, so $I(T, X) \approx 0$ and $I(T, Y) \approx 0$. 
Next, as LLM gradually fits the pre-training dataset, its abilities in language understanding and concept modeling enhance, contributing to increases in both $I(T, X)$ and $I(T, Y)$. 
As pre-training progresses, LLM learns to better compress the irrelevant information in the dataset and preserve more label-related information (i.e., trustworthiness), leading to a reduction in $I(T, X)$ and an improvement in $I(T, Y)$.
Overall, we are at the forefront of investigating the phase transition from ``fitting'' to ``compression'' in the context of trustworthiness during pre-training. We hope that our insights will motivate further exploration of LLMs' pre-training dynamics.

\section{Related Work}

\paragraph{Probing LLM representations.} 

Probing classifiers~\citep{alain2016understanding,tenney2019you,pimentel2020information,li2021implicit,belinkov2022probing,rauker2023toward} is one of the prominent methods for identifying certain properties acquired by the language model~\citep{zhao2023explainability}. 
% Equipped with this probing tool, 
Researchers probe LLMs and discover linear separable patterns within LLMs, including space and time~\citep{gurnee2023language}, game states~\citep{nanda2023emergent},  answerability~\citep{slobodkin2023curious}, and some counterfactual pairs of concepts~\citep{park2023linear}. 
It is also observed that LLM representations contain linearly separable patterns about trustworthiness, such as truthfulness~\citep{li2023inferencetime,marks2023geometry,zou2023representation}. However, they do not probe LLM representations during pre-training. In this work, we consider the whole pre-training period of LLMs and probe their presentations dynamically.

\paragraph{Steering vectors for trustworthy LLMs.}
Numerous intriguing approaches have been proposed to investigate the trustworthiness of LLMs~\cite{ouyang2022training,rafailov2024direct,zhang2024psysafe,li2024salad,ren2024exploring}. Specifically, some promising approaches explore the latent space, utilizing representations to improve model performance~\cite{liu2023context,jorgensen2023improving}. Various studies investigate activation engineering within LLMs from both theoretical and practical perspectives, affecting model performance by manipulating the model's representational space~\cite{park2023linear,turner2023activation,zou2023representation}. Furthermore, \citet{wang2023backdoor}, \citet{rimsky2023steering} and \citet{wang2024inferaligner} construct directional vectors to explore the model's safety and alignment, with the goal of making models helpful, honest, and harmless. However, there has been no investigation into how representations change during the pre-training phase of LLMs. In this paper, we explore and leverage representations during this phase, paving the way for new research avenues in activation engineering.

\paragraph{Understanding the training process of DNNs.} 

Many empirical studies observe that DNNs tend to learn simple concepts during the learning process \cite{arpit2017closer, liu2021impact, mangalam2019deep}. Furthermore, \citet{xu2019frequency}, \citet{liu2023towards}, \citet{zhou2024explaining}, and \citet{tian2023joma} theoretically explain the learning preference of DNNs. Meanwhile, many researchers focus on analyzing the utility of fine-tuning for language models \cite{merchant2020happens, hao-etal-2020-investigating, aghajanyan2021intrinsic, zhou2022closer, mosbach2020interplay} and attempt to understand the in-context learning \cite{ren2024identifying}. However, few previous studies investigate how trustworthiness is learned by LLMs during pre-training. In this paper, we take a closer look at the learning dynamic of trustworthiness within LLMs' representations.

\section{Discussion}
% self-alignment (from jiqizhixin)
As the capabilities of LLMs have increased, conventional alignment techniques that rely on ``human feedback'' (like RLHF) may no longer work when trying to align models that are more powerful than humans~\citep{burns2023weak,yuan2024self}. 
To address this challenge, research institutions are actively exploring new solutions. For example, OpenAI introduces ``superalignment''~\footnote{https://openai.com/index/introducing-superalignment/} and proposes a ``weak-to-strong supervision'' approach~\citep{burns2023weak}. Also, Meta proposes a ``self-reward'' mechanism~\citep{yuan2024self}. At the same time, more and more research focuses on the emerging field of ``self-alignment''~\citep{sun2023principle,li2023self}.
In this paper, we provide a deeper understanding of the pre-training dynamics and successfully align the SFT model using its own pre-training checkpoints. We believe that the pre-training period is worth being explored and it may be a promising source for self-alignment.

% resource-conserving (from magic)
On the other hand, to make LLMs trustworthy, recent conventional alignment methods, such as SFT and RLHF, incur high costs due to exhaustive human annotations~\citep{wang2022self,honovich2022unnatural,sun2023principle} and time-consuming instruction tuning~\citep{zhou2023lima,chen2023maybe,chen2023alpagasus}. 
In this paper, we delve into the pre-training period to enhance trustworthiness without collecting data or tuning the model.
We expect more alignment approaches inspired by the pre-training phase (like~\citet{korbak2023pretraining}) and to circumvent potential alignment costs in the future.

\section{Conclusion}
In this work, we take an initial and illuminating step towards elucidating the conceptual understanding of trustworthiness during pre-training.
Firstly, by linear probing LLMs across reliability, privacy, toxicity, fairness, and robustness, we investigate the ability of LLMs representations to discern opposing concepts within each trustworthiness dimension during the whole pre-training period. 
Furthermore, motivated by the probing results, we conduct extensive experiments to reveal the potential of utilizing representations from LLMs during its previous pre-training period to enhance LLMs' own trustworthiness. 
Finally, we use mutual information to probe LLMs during pre-training and reveal some similarities in the learning mechanism between LLMs and traditional DNNs.
% conclusion
Taken collectively, the empirical study presented in this work can not only justify the potential to improve the trustworthiness of LLMs using their own pre-training checkpoints but may also lead to a better understanding of the dynamics of LLM representations, especially the trustworthiness-related concepts.

\section*{Acknowledgements}
We thank the anonymous reviewers for their constructive suggestions to improve the quality of this paper. This work is supported by the Beijing Natural Science Foundation (No.4222029); the National Natural Science Foundation of China (NO.62076234); the National Key Research and Development Project (No.2022YFB2703102); the “Intelligent Social Governance Interdisciplinary Platform, Major Innovation \& Planning Interdisciplinary Platform for the “Double-First Class” Initiative, Renmin University of China”; the Beijing Outstanding Young Scientist Program (NO.BJJWZYJH012019100020098); the Public Computing Cloud, Renmin University of China; the Fundamental Research Funds for the Central Universities, and the Research Funds of Renmin University of China (NO.2021030199); the Huawei-Renmin University joint program on Information Retrieval; and the Unicom Innovation Ecological Cooperation Plan.

\section*{Limitations}

There are several limitations of this work. 
Firstly, we only focus on five essential trustworthiness dimensions and do not encompass all the dimensions, such as those that appeared in~\cite {doi/10.2759/346720,Liu2023trustworthy_ai}.
A wide variety of definitions for each trustworthiness dimension, as discussed by~\cite{wang2024decodingtrust,sun2024trustllm}, are not completely covered in our analysis.
Secondly, 
due to the absence of publicly available larger and more complex LLMs (such as 13B or others) that release pre-training period checkpoints, we are limited to conducting experiments on 7B 
series models (we also provide some experimental results on another 7B-size model named OLMo~\citep{groeneveld2024olmo} in Appendix.\ref{appendix-olmo}).
Finally, 
for evaluation of TruthfulQA, the precision of evaluation results depends on the performance of the ``GPT-judge'' evaluator. And for multiple-choice evaluation, the evaluation results may rely on the instruction following ability of LLMs.

\section*{Broader Impact and Ethics Statement}

This study concentrates on better understanding the learning dynamics of LLM trustworthiness during pre-training. The motivation of our steering vector experiments is centered on improving the trustworthiness of LLMs. We recognize the sensitive nature of our research and ensure that it strictly complies with legal and ethical guidelines.

This research is carried out in a secure, controlled environment, ensuring the safety of real-world systems. 
Given the nature of our work, which includes dealing with potentially sensitive content like unreliable statements and toxic sentences, we have implemented strict protocols. 
Access to the most sensitive aspects of our experiments is limited to researchers with the proper authorization, who are committed to following rigorous ethical standards. 
These precautions are taken to maintain the integrity of our research and to mitigate any risks that could arise from the experiment's content.

\bibliography{custom}

\begin{thebibliography}{106}
\expandafter\ifx\csname natexlab\endcsname\relax\def\natexlab#1{#1}\fi

\bibitem[{Aghajanyan et~al.(2021)Aghajanyan, Gupta, and Zettlemoyer}]{aghajanyan2021intrinsic}
Armen Aghajanyan, Sonal Gupta, and Luke Zettlemoyer. 2021.
\newblock Intrinsic dimensionality explains the effectiveness of language model fine-tuning.
\newblock In \emph{Proceedings of the 59th Annual Meeting of the Association for Computational Linguistics and the 11th International Joint Conference on Natural Language Processing (Volume 1: Long Papers)}, pages 7319--7328.

\bibitem[{Alain and Bengio(2016)}]{alain2016understanding}
Guillaume Alain and Yoshua Bengio. 2016.
\newblock Understanding intermediate layers using linear classifier probes.
\newblock \emph{arXiv preprint arXiv:1610.01644}.

\bibitem[{Alemi et~al.(2016)Alemi, Fischer, Dillon, and Murphy}]{alemi2016deep}
Alexander~A Alemi, Ian Fischer, Joshua~V Dillon, and Kevin Murphy. 2016.
\newblock Deep variational information bottleneck.
\newblock \emph{arXiv preprint arXiv:1612.00410}.

\bibitem[{Arpit et~al.(2017)Arpit, Jastrz{\k{e}}bski, Ballas, Krueger, Bengio, Kanwal, Maharaj, Fischer, Courville, Bengio et~al.}]{arpit2017closer}
Devansh Arpit, Stanis{\l}aw Jastrz{\k{e}}bski, Nicolas Ballas, David Krueger, Emmanuel Bengio, Maxinder~S Kanwal, Tegan Maharaj, Asja Fischer, Aaron Courville, Yoshua Bengio, et~al. 2017.
\newblock A closer look at memorization in deep networks.
\newblock In \emph{International conference on machine learning}, pages 233--242. PMLR.

\bibitem[{Azaria and Mitchell(2023)}]{azaria2023internal}
Amos Azaria and Tom Mitchell. 2023.
\newblock The internal state of an llm knows when its lying.
\newblock \emph{arXiv preprint arXiv:2304.13734}.

\bibitem[{Belinkov(2022)}]{belinkov2022probing}
Yonatan Belinkov. 2022.
\newblock Probing classifiers: Promises, shortcomings, and advances.
\newblock \emph{Computational Linguistics}, 48(1):207--219.

\bibitem[{Burns et~al.(2023)Burns, Izmailov, Kirchner, Baker, Gao, Aschenbrenner, Chen, Ecoffet, Joglekar, Leike et~al.}]{burns2023weak}
Collin Burns, Pavel Izmailov, Jan~Hendrik Kirchner, Bowen Baker, Leo Gao, Leopold Aschenbrenner, Yining Chen, Adrien Ecoffet, Manas Joglekar, Jan Leike, et~al. 2023.
\newblock Weak-to-strong generalization: Eliciting strong capabilities with weak supervision.
\newblock \emph{arXiv preprint arXiv:2312.09390}.

\bibitem[{Burns et~al.(2022)Burns, Ye, Klein, and Steinhardt}]{burns2022discovering}
Collin Burns, Haotian Ye, Dan Klein, and Jacob Steinhardt. 2022.
\newblock Discovering latent knowledge in language models without supervision.
\newblock \emph{arXiv preprint arXiv:2212.03827}.

\bibitem[{Chen et~al.(2023{\natexlab{a}})Chen, Zhang, Zhang, Yang, Hu, Ma, Yanggong, and Zhao}]{chen2023maybe}
Hao Chen, Yiming Zhang, Qi~Zhang, Hantao Yang, Xiaomeng Hu, Xuetao Ma, Yifan Yanggong, and Junbo Zhao. 2023{\natexlab{a}}.
\newblock Maybe only 0.5\% data is needed: A preliminary exploration of low training data instruction tuning.
\newblock \emph{arXiv preprint arXiv:2305.09246}.

\bibitem[{Chen et~al.(2024)Chen, Wang, Wang, Chen, and Wang}]{chen2024taichi}
Huimin Chen, Chengyu Wang, Yanhao Wang, Cen Chen, and Yinggui Wang. 2024.
\newblock Taichi: Improving the robustness of nlp models by seeking common ground while reserving differences.
\newblock In \emph{Proceedings of the 2024 Joint International Conference on Computational Linguistics, Language Resources and Evaluation (LREC-COLING 2024)}, pages 15542--15551.

\bibitem[{Chen et~al.(2023{\natexlab{b}})Chen, Li, Yan, Wang, Gunaratna, Yadav, Tang, Srinivasan, Zhou, Huang et~al.}]{chen2023alpagasus}
Lichang Chen, Shiyang Li, Jun Yan, Hai Wang, Kalpa Gunaratna, Vikas Yadav, Zheng Tang, Vijay Srinivasan, Tianyi Zhou, Heng Huang, et~al. 2023{\natexlab{b}}.
\newblock Alpagasus: Training a better alpaca with fewer data.
\newblock \emph{arXiv preprint arXiv:2307.08701}.

\bibitem[{Choi et~al.(2023)Choi, Jung, and Watanabe}]{choi2023understanding}
Kwanghee Choi, Jee-weon Jung, and Shinji Watanabe. 2023.
\newblock Understanding probe behaviors through variational bounds of mutual information.
\newblock \emph{arXiv preprint arXiv:2312.10019}.

\bibitem[{Commission(2021b)}]{AI_Act}
European Commission. 2021b.
\newblock Proposal for a regulation of the european parliament and of the council laying down harmonised rules on artificial intelligence (artificial intelligence act) and amending certain union legislative acts, pub. l. no. com(2021) 206 final.

\bibitem[{Commission et~al.(2019)Commission, Directorate-General~for Communications~Networks, and Technology}]{doi/10.2759/346720}
European Commission, Content Directorate-General~for Communications~Networks, and Technology. 2019.
\newblock \href {https://doi.org/doi/10.2759/346720} {\emph{Ethics guidelines for trustworthy AI}}.
\newblock Publications Office.

\bibitem[{Cover(1999)}]{cover1999elements}
Thomas~M Cover. 1999.
\newblock \emph{Elements of information theory}.
\newblock John Wiley \& Sons.

\bibitem[{Foundation(2023)}]{AI_Verify}
AI~Verify Foundation. 2023.
\newblock \href {https://aiverifyfoundation.sg/downloads/Cataloguing_LLM_Evaluations.pdf} {Catalogue of llm evaluations}.

\bibitem[{Gao et~al.(2023)Gao, Tow, Abbasi, Biderman, Black, DiPofi, Foster, Golding, Hsu, Le~Noac'h, Li, McDonell, Muennighoff, Ociepa, Phang, Reynolds, Schoelkopf, Skowron, Sutawika, Tang, Thite, Wang, Wang, and Zou}]{eval-harness}
Leo Gao, Jonathan Tow, Baber Abbasi, Stella Biderman, Sid Black, Anthony DiPofi, Charles Foster, Laurence Golding, Jeffrey Hsu, Alain Le~Noac'h, Haonan Li, Kyle McDonell, Niklas Muennighoff, Chris Ociepa, Jason Phang, Laria Reynolds, Hailey Schoelkopf, Aviya Skowron, Lintang Sutawika, Eric Tang, Anish Thite, Ben Wang, Kevin Wang, and Andy Zou. 2023.
\newblock A framework for few-shot language model evaluation.

\bibitem[{Geiger(2021)}]{geiger2021information}
Bernhard~C Geiger. 2021.
\newblock On information plane analyses of neural network classifiers--a review.
\newblock \emph{IEEE Transactions on Neural Networks and Learning Systems}.

\bibitem[{Goldfeld and Polyanskiy(2020)}]{goldfeld2020information}
Ziv Goldfeld and Yury Polyanskiy. 2020.
\newblock The information bottleneck problem and its applications in machine learning.
\newblock \emph{IEEE Journal on Selected Areas in Information Theory}, 1(1):19--38.

\bibitem[{Gretton et~al.(2005)Gretton, Bousquet, Smola, and Sch{\"o}lkopf}]{gretton2005measuring}
Arthur Gretton, Olivier Bousquet, Alex Smola, and Bernhard Sch{\"o}lkopf. 2005.
\newblock Measuring statistical dependence with hilbert-schmidt norms.
\newblock In \emph{International conference on algorithmic learning theory}, pages 63--77.

\bibitem[{Groeneveld et~al.(2024)Groeneveld, Beltagy, Walsh, Bhagia, Kinney, Tafjord, Jha, Ivison, Magnusson, Wang et~al.}]{groeneveld2024olmo}
Dirk Groeneveld, Iz~Beltagy, Pete Walsh, Akshita Bhagia, Rodney Kinney, Oyvind Tafjord, Ananya~Harsh Jha, Hamish Ivison, Ian Magnusson, Yizhong Wang, et~al. 2024.
\newblock Olmo: Accelerating the science of language models.
\newblock \emph{arXiv preprint arXiv:2402.00838}.

\bibitem[{Gurnee and Tegmark(2023)}]{gurnee2023language}
Wes Gurnee and Max Tegmark. 2023.
\newblock Language models represent space and time.
\newblock \emph{arXiv preprint arXiv:2310.02207}.

\bibitem[{Hao et~al.(2020)Hao, Dong, Wei, and Xu}]{hao-etal-2020-investigating}
Yaru Hao, Li~Dong, Furu Wei, and Ke~Xu. 2020.
\newblock \href {https://aclanthology.org/2020.aacl-main.11} {Investigating learning dynamics of {BERT} fine-tuning}.
\newblock In \emph{Proceedings of the 1st Conference of the Asia-Pacific Chapter of the Association for Computational Linguistics and the 10th International Joint Conference on Natural Language Processing}, pages 87--92. Association for Computational Linguistics.

\bibitem[{Hartvigsen et~al.(2022)Hartvigsen, Gabriel, Palangi, Sap, Ray, and Kamar}]{hartvigsen2022toxigen}
Thomas Hartvigsen, Saadia Gabriel, Hamid Palangi, Maarten Sap, Dipankar Ray, and Ece Kamar. 2022.
\newblock Toxigen: A large-scale machine-generated dataset for implicit and adversarial hate speech detection.
\newblock In \emph{Proceedings of the 60th Annual Meeting of the Association for Computational Linguistics}.

\bibitem[{Hayes(2020)}]{robustness_privacy_1}
Jamie Hayes. 2020.
\newblock Trade-offs between membership privacy \& adversarially robust learning.
\newblock \emph{arXiv preprint arXiv:2006.04622}.

\bibitem[{Hewitt and Manning(2019)}]{hewitt2019structural}
John Hewitt and Christopher~D Manning. 2019.
\newblock A structural probe for finding syntax in word representations.
\newblock In \emph{Proceedings of the 2019 Conference of the North American Chapter of the Association for Computational Linguistics: Human Language Technologies, Volume 1 (Long and Short Papers)}, pages 4129--4138.

\bibitem[{Honovich et~al.(2022)Honovich, Scialom, Levy, and Schick}]{honovich2022unnatural}
Or~Honovich, Thomas Scialom, Omer Levy, and Timo Schick. 2022.
\newblock Unnatural instructions: Tuning language models with (almost) no human labor.
\newblock \emph{arXiv preprint arXiv:2212.09689}.

\bibitem[{Hosseini et~al.(2023)Hosseini, Palangi, and Awadallah}]{hosseini-etal-2023-empirical}
Saghar Hosseini, Hamid Palangi, and Ahmed~Hassan Awadallah. 2023.
\newblock An empirical study of metrics to measure representational harms in pre-trained language models.
\newblock In \emph{Proceedings of the 3rd Workshop on Trustworthy Natural Language Processing (TrustNLP 2023)}, pages 121--134.

\bibitem[{Ji et~al.(2023)Ji, Liu, Dai, Pan, Zhang, Bian, Chen, Sun, Wang, and Yang}]{ji2023beavertails}
Jiaming Ji, Mickel Liu, Juntao Dai, Xuehai Pan, Chi Zhang, Ce~Bian, Boyuan Chen, Ruiyang Sun, Yizhou Wang, and Yaodong Yang. 2023.
\newblock Beavertails: Towards improved safety alignment of {LLM} via a human-preference dataset.
\newblock In \emph{Thirty-seventh Conference on Neural Information Processing Systems Datasets and Benchmarks Track}.

\bibitem[{Jorgensen et~al.(2023)Jorgensen, Cope, Schoots, and Shanahan}]{jorgensen2023improving}
Ole Jorgensen, Dylan Cope, Nandi Schoots, and Murray Shanahan. 2023.
\newblock Improving activation steering in language models with mean-centring.
\newblock \emph{arXiv preprint arXiv:2312.03813}.

\bibitem[{Korbak et~al.(2023)Korbak, Shi, Chen, Bhalerao, Buckley, Phang, Bowman, and Perez}]{korbak2023pretraining}
Tomasz Korbak, Kejian Shi, Angelica Chen, Rasika~Vinayak Bhalerao, Christopher Buckley, Jason Phang, Samuel~R Bowman, and Ethan Perez. 2023.
\newblock Pretraining language models with human preferences.
\newblock In \emph{International Conference on Machine Learning}, pages 17506--17533.

\bibitem[{Kraskov et~al.(2004)Kraskov, St{\"o}gbauer, and Grassberger}]{kraskov2004estimating}
Alexander Kraskov, Harald St{\"o}gbauer, and Peter Grassberger. 2004.
\newblock Estimating mutual information.
\newblock \emph{Physical review E}, 69(6):066138.

\bibitem[{Li et~al.(2021)Li, Nye, and Andreas}]{li2021implicit}
Belinda~Z Li, Maxwell Nye, and Jacob Andreas. 2021.
\newblock Implicit representations of meaning in neural language models.
\newblock \emph{arXiv preprint arXiv:2106.00737}.

\bibitem[{Li et~al.(2023{\natexlab{a}})Li, Patel, Vi{\'e}gas, Pfister, and Wattenberg}]{li2023inferencetime}
Kenneth Li, Oam Patel, Fernanda Vi{\'e}gas, Hanspeter Pfister, and Martin Wattenberg. 2023{\natexlab{a}}.
\newblock Inference-time intervention: Eliciting truthful answers from a language model.
\newblock In \emph{Thirty-seventh Conference on Neural Information Processing Systems}.

\bibitem[{Li et~al.(2024)Li, Dong, Wang, Hu, Zuo, Lin, Qiao, and Shao}]{li2024salad}
Lijun Li, Bowen Dong, Ruohui Wang, Xuhao Hu, Wangmeng Zuo, Dahua Lin, Yu~Qiao, and Jing Shao. 2024.
\newblock Salad-bench: A hierarchical and comprehensive safety benchmark for large language models.
\newblock \emph{arXiv preprint arXiv:2402.05044}.

\bibitem[{Li et~al.(2023{\natexlab{b}})Li, Yu, Zhou, Schick, Zettlemoyer, Levy, Weston, and Lewis}]{li2023self}
Xian Li, Ping Yu, Chunting Zhou, Timo Schick, Luke Zettlemoyer, Omer Levy, Jason Weston, and Mike Lewis. 2023{\natexlab{b}}.
\newblock Self-alignment with instruction backtranslation.
\newblock \emph{arXiv preprint arXiv:2308.06259}.

\bibitem[{Liang et~al.(2022)Liang, Bommasani, Lee, Tsipras, Soylu, Yasunaga, Zhang, Narayanan, Wu, Kumar et~al.}]{liang2022holistic}
Percy Liang, Rishi Bommasani, Tony Lee, Dimitris Tsipras, Dilara Soylu, Michihiro Yasunaga, Yian Zhang, Deepak Narayanan, Yuhuai Wu, Ananya Kumar, et~al. 2022.
\newblock Holistic evaluation of language models.
\newblock \emph{arXiv preprint arXiv:2211.09110}.

\bibitem[{Lin et~al.(2022)Lin, Hilton, and Evans}]{Lin_Hilton_Evans_2022}
Stephanie Lin, Jacob Hilton, and Owain Evans. 2022.
\newblock \href {https://doi.org/10.18653/v1/2022.acl-long.229} {Truthfulqa: Measuring how models mimic human falsehoods}.
\newblock In \emph{Proceedings of the 60th Annual Meeting of the Association for Computational Linguistics (Volume 1: Long Papers)}.

\bibitem[{Liu et~al.(2024)Liu, Han, Wang, Tsvetkov, Choi, and Smith}]{liu2024tuning}
Alisa Liu, Xiaochuang Han, Yizhong Wang, Yulia Tsvetkov, Yejin Choi, and Noah~A. Smith. 2024.
\newblock \href {http://arxiv.org/abs/2401.08565} {Tuning language models by proxy}.

\bibitem[{Liu et~al.(2021)Liu, Huang, Salzmann, Zhang, and S{\"u}sstrunk}]{liu2021impact}
Chen Liu, Zhichao Huang, Mathieu Salzmann, Tong Zhang, and Sabine S{\"u}sstrunk. 2021.
\newblock On the impact of hard adversarial instances on overfitting in adversarial training.
\newblock \emph{arXiv preprint arXiv:2112.07324}.

\bibitem[{Liu et~al.(2023{\natexlab{a}})Liu, Deng, Cheng, Ren, Wang, and Zhang}]{liu2023towards}
Dongrui Liu, Huiqi Deng, Xu~Cheng, Qihan Ren, Kangrui Wang, and Quanshi Zhang. 2023{\natexlab{a}}.
\newblock Towards the difficulty for a deep neural network to learn concepts of different complexities.
\newblock In \emph{Thirty-seventh Conference on Neural Information Processing Systems}.

\bibitem[{Liu et~al.(2023{\natexlab{b}})Liu, Wang, Fan, Liu, Li, Jain, Liu, Jain, and Tang}]{Liu2023trustworthy_ai}
Haochen Liu, Yiqi Wang, Wenqi Fan, Xiaorui Liu, Yaxin Li, Shaili Jain, Yunhao Liu, Anil Jain, and Jiliang Tang. 2023{\natexlab{b}}.
\newblock \href {https://doi.org/10.1145/3546872} {Trustworthy ai: A computational perspective}.
\newblock \emph{ACM Transactions on Intelligent Systems and Technology}, page 1–59.

\bibitem[{Liu et~al.(2023{\natexlab{c}})Liu, Xing, and Zou}]{liu2023context}
Sheng Liu, Lei Xing, and James Zou. 2023{\natexlab{c}}.
\newblock In-context vectors: Making in context learning more effective and controllable through latent space steering.
\newblock \emph{arXiv preprint arXiv:2311.06668}.

\bibitem[{Liu et~al.(2023{\natexlab{d}})Liu, Yao, Ton, Zhang, Guo, Cheng, Klochkov, Taufiq, and Li}]{liu2023trustworthy}
Yang Liu, Yuanshun Yao, Jean-Francois Ton, Xiaoying Zhang, Ruocheng Guo, Hao Cheng, Yegor Klochkov, Muhammad~Faaiz Taufiq, and Hang Li. 2023{\natexlab{d}}.
\newblock \href {http://arxiv.org/abs/2308.05374} {Trustworthy llms: a survey and guideline for evaluating large language models' alignment}.

\bibitem[{Liu et~al.(2020)Liu, Ott, Goyal, Du, Joshi, Chen, Levy, Lewis, Zettlemoyer, and Stoyanov}]{liu2020roberta}
Yinhan Liu, Myle Ott, Naman Goyal, Jingfei Du, Mandar Joshi, Danqi Chen, Omer Levy, Mike Lewis, Luke Zettlemoyer, and Veselin Stoyanov. 2020.
\newblock Ro{\{}bert{\}}a: A robustly optimized {\{}bert{\}} pretraining approach.

\bibitem[{Liu et~al.(2023{\natexlab{e}})Liu, Qiao, Neiswanger, Wang, Tan, Tao, Li, Wang, Sun, Pangarkar et~al.}]{liu2023llm360}
Zhengzhong Liu, Aurick Qiao, Willie Neiswanger, Hongyi Wang, Bowen Tan, Tianhua Tao, Junbo Li, Yuqi Wang, Suqi Sun, Omkar Pangarkar, et~al. 2023{\natexlab{e}}.
\newblock Llm360: Towards fully transparent open-source llms.
\newblock \emph{arXiv preprint arXiv:2312.06550}.

\bibitem[{Lorenzen et~al.(2021)Lorenzen, Igel, and Nielsen}]{lorenzen2021information}
Stephan~Sloth Lorenzen, Christian Igel, and Mads Nielsen. 2021.
\newblock Information bottleneck: Exact analysis of (quantized) neural networks.
\newblock \emph{arXiv preprint arXiv:2106.12912}.

\bibitem[{Ma et~al.(2020)Ma, Lewis, and Kleijn}]{ma2020hsic}
Wan-Duo~Kurt Ma, JP~Lewis, and W~Bastiaan Kleijn. 2020.
\newblock The hsic bottleneck: Deep learning without back-propagation.
\newblock In \emph{Proceedings of the AAAI conference on artificial intelligence}, pages 5085--5092.

\bibitem[{Mangalam and Prabhu(2019)}]{mangalam2019deep}
Karttikeya Mangalam and Vinay~Uday Prabhu. 2019.
\newblock Do deep neural networks learn shallow learnable examples first?

\bibitem[{Mangold et~al.(2023)Mangold, Perrot, Bellet, and Tommasi}]{privacy_fairness_1}
Paul Mangold, Micha{\"e}l Perrot, Aur{\'e}lien Bellet, and Marc Tommasi. 2023.
\newblock Differential privacy has bounded impact on fairness in classification.
\newblock In \emph{International Conference on Machine Learning}, pages 23681--23705.

\bibitem[{Marks and Tegmark(2023)}]{marks2023geometry}
Samuel Marks and Max Tegmark. 2023.
\newblock The geometry of truth: Emergent linear structure in large language model representations of true/false datasets.
\newblock \emph{arXiv preprint arXiv:2310.06824}.

\bibitem[{Merchant et~al.(2020)Merchant, Rahimtoroghi, Pavlick, and Tenney}]{merchant2020happens}
Amil Merchant, Elahe Rahimtoroghi, Ellie Pavlick, and Ian Tenney. 2020.
\newblock What happens to bert embeddings during fine-tuning?
\newblock In \emph{Proceedings of the Third BlackboxNLP Workshop on Analyzing and Interpreting Neural Networks for NLP}, pages 33--44.

\bibitem[{Mireshghallah et~al.(2023)Mireshghallah, Kim, Zhou, Tsvetkov, Sap, Shokri, and Choi}]{mireshghallah2023llms}
Niloofar Mireshghallah, Hyunwoo Kim, Xuhui Zhou, Yulia Tsvetkov, Maarten Sap, Reza Shokri, and Yejin Choi. 2023.
\newblock \href {http://arxiv.org/abs/2310.17884} {Can llms keep a secret? testing privacy implications of language models via contextual integrity theory}.

\bibitem[{Mitchell et~al.(2023)Mitchell, Rafailov, Sharma, Finn, and Manning}]{mitchell2023emulator}
Eric Mitchell, Rafael Rafailov, Archit Sharma, Chelsea Finn, and Christopher~D. Manning. 2023.
\newblock \href {http://arxiv.org/abs/2310.12962} {An emulator for fine-tuning large language models using small language models}.

\bibitem[{Mosbach et~al.(2020)Mosbach, Khokhlova, Hedderich, and Klakow}]{mosbach2020interplay}
Marius Mosbach, Anna Khokhlova, Michael~A Hedderich, and Dietrich Klakow. 2020.
\newblock On the interplay between fine-tuning and sentence-level probing for linguistic knowledge in pre-trained transformers.
\newblock In \emph{Findings of the Association for Computational Linguistics: EMNLP 2020}, pages 2502--2516.

\bibitem[{Nadeem et~al.(2021)Nadeem, Bethke, and Reddy}]{stereoset}
Moin Nadeem, Anna Bethke, and Siva Reddy. 2021.
\newblock {S}tereo{S}et: Measuring stereotypical bias in pretrained language models.
\newblock In \emph{Proceedings of the 59th Annual Meeting of the Association for Computational Linguistics and the 11th International Joint Conference on Natural Language Processing (Volume 1: Long Papers)}, pages 5356--5371.

\bibitem[{Nanda et~al.(2023)Nanda, Lee, and Wattenberg}]{nanda2023emergent}
Neel Nanda, Andrew Lee, and Martin Wattenberg. 2023.
\newblock Emergent linear representations in world models of self-supervised sequence models.
\newblock \emph{arXiv preprint arXiv:2309.00941}.

\bibitem[{Newman(2023)}]{CLTC}
Jessica Newman. 2023.
\newblock \href {https://cltc.berkeley.edu/publication/a-taxonomy-of-trustworthiness-for-artificial-intelligence/} {A taxonomy of trustworthiness for artificial intelligence: Connecting properties of trustworthiness with risk management and the ai lifecycle}.

\bibitem[{Noshad et~al.(2019)Noshad, Zeng, and Hero}]{noshad2019scalable}
Morteza Noshad, Yu~Zeng, and Alfred~O Hero. 2019.
\newblock Scalable mutual information estimation using dependence graphs.
\newblock In \emph{ICASSP 2019-2019 IEEE International Conference on Acoustics, Speech and Signal Processing (ICASSP)}, pages 2962--2966. IEEE.

\bibitem[{Ouyang et~al.(2022)Ouyang, Wu, Jiang, Almeida, Wainwright, Mishkin, Zhang, Agarwal, Slama, Ray et~al.}]{ouyang2022training}
Long Ouyang, Jeffrey Wu, Xu~Jiang, Diogo Almeida, Carroll Wainwright, Pamela Mishkin, Chong Zhang, Sandhini Agarwal, Katarina Slama, Alex Ray, et~al. 2022.
\newblock Training language models to follow instructions with human feedback.
\newblock \emph{Advances in Neural Information Processing Systems}, 35:27730--27744.

\bibitem[{Paperno et~al.(2016)Paperno, Kruszewski, Lazaridou, Pham, Bernardi, Pezzelle, Baroni, Boleda, and Fern{\'a}ndez}]{paperno-etal-2016-lambada}
Denis Paperno, Germ{\'a}n Kruszewski, Angeliki Lazaridou, Ngoc~Quan Pham, Raffaella Bernardi, Sandro Pezzelle, Marco Baroni, Gemma Boleda, and Raquel Fern{\'a}ndez. 2016.
\newblock The {LAMBADA} dataset: Word prediction requiring a broad discourse context.
\newblock In \emph{Proceedings of the 54th Annual Meeting of the Association for Computational Linguistics (Volume 1: Long Papers)}, pages 1525--1534.

\bibitem[{Park et~al.(2023)Park, Choe, and Veitch}]{park2023linear}
Kiho Park, Yo~Joong Choe, and Victor Veitch. 2023.
\newblock The linear representation hypothesis and the geometry of large language models.
\newblock \emph{arXiv preprint arXiv:2311.03658}.

\bibitem[{Pimentel et~al.(2020)Pimentel, Valvoda, Maudslay, Zmigrod, Williams, and Cotterell}]{pimentel2020information}
Tiago Pimentel, Josef Valvoda, Rowan~Hall Maudslay, Ran Zmigrod, Adina Williams, and Ryan Cotterell. 2020.
\newblock \href {https://doi.org/10.18653/v1/2020.acl-main.420} {Information-theoretic probing for linguistic structure}.
\newblock In \emph{Proceedings of the 58th Annual Meeting of the Association for Computational Linguistics}, pages 4609--4622.

\bibitem[{Poole et~al.(2019)Poole, Ozair, Van Den~Oord, Alemi, and Tucker}]{poole2019variational}
Ben Poole, Sherjil Ozair, Aaron Van Den~Oord, Alex Alemi, and George Tucker. 2019.
\newblock On variational bounds of mutual information.
\newblock In \emph{International Conference on Machine Learning}, pages 5171--5180.

\bibitem[{Radford et~al.(2019)Radford, Wu, Child, Luan, Amodei, and Sutskever}]{radford2019language}
Alec Radford, Jeff Wu, Rewon Child, David Luan, Dario Amodei, and Ilya Sutskever. 2019.
\newblock Language models are unsupervised multitask learners.

\bibitem[{Rafailov et~al.(2024)Rafailov, Sharma, Mitchell, Manning, Ermon, and Finn}]{rafailov2024direct}
Rafael Rafailov, Archit Sharma, Eric Mitchell, Christopher~D Manning, Stefano Ermon, and Chelsea Finn. 2024.
\newblock Direct preference optimization: Your language model is secretly a reward model.
\newblock \emph{Advances in Neural Information Processing Systems}, 36.

\bibitem[{R{\"a}uker et~al.(2023)R{\"a}uker, Ho, Casper, and Hadfield-Menell}]{rauker2023toward}
Tilman R{\"a}uker, Anson Ho, Stephen Casper, and Dylan Hadfield-Menell. 2023.
\newblock Toward transparent ai: A survey on interpreting the inner structures of deep neural networks.
\newblock In \emph{2023 IEEE Conference on Secure and Trustworthy Machine Learning (SaTML)}, pages 464--483. IEEE.

\bibitem[{Ravichander et~al.(2021)Ravichander, Belinkov, and Hovy}]{Ravichander_Belinkov_Hovy_2020}
Abhilasha Ravichander, Yonatan Belinkov, and Eduard Hovy. 2021.
\newblock \href {https://doi.org/10.18653/v1/2021.eacl-main.295} {Probing the probing paradigm: Does probing accuracy entail task relevance?}
\newblock In \emph{Proceedings of the 16th Conference of the European Chapter of the Association for Computational Linguistics: Main Volume}, pages 3363--3377.

\bibitem[{Ren et~al.(2024{\natexlab{a}})Ren, Guo, Yan, Liu, Qiu, and Lin}]{ren2024identifying}
Jie Ren, Qipeng Guo, Hang Yan, Dongrui Liu, Xipeng Qiu, and Dahua Lin. 2024{\natexlab{a}}.
\newblock Identifying semantic induction heads to understand in-context learning.
\newblock \emph{arXiv preprint arXiv:2402.13055}.

\bibitem[{Ren et~al.(2024{\natexlab{b}})Ren, Gao, Shao, Yan, Tan, Lam, and Ma}]{ren2024exploring}
Qibing Ren, Chang Gao, Jing Shao, Junchi Yan, Xin Tan, Wai Lam, and Lizhuang Ma. 2024{\natexlab{b}}.
\newblock Exploring safety generalization challenges of large language models via code.
\newblock \emph{arXiv preprint arXiv:2403.07865}.

\bibitem[{Rimsky et~al.(2023)Rimsky, Gabrieli, Schulz, Tong, Hubinger, and Turner}]{rimsky2023steering}
Nina Rimsky, Nick Gabrieli, Julian Schulz, Meg Tong, Evan Hubinger, and Alexander~Matt Turner. 2023.
\newblock Steering llama 2 via contrastive activation addition.
\newblock \emph{arXiv preprint arXiv:2312.06681}.

\bibitem[{Saxe et~al.(2019)Saxe, Bansal, Dapello, Advani, Kolchinsky, Tracey, and Cox}]{saxe2019information}
Andrew~M Saxe, Yamini Bansal, Joel Dapello, Madhu Advani, Artemy Kolchinsky, Brendan~D Tracey, and David~D Cox. 2019.
\newblock On the information bottleneck theory of deep learning.
\newblock \emph{Journal of Statistical Mechanics: Theory and Experiment}, 2019(12):124020.

\bibitem[{Shwartz-Ziv and Tishby(2017)}]{shwartz2017opening}
Ravid Shwartz-Ziv and Naftali Tishby. 2017.
\newblock Opening the black box of deep neural networks via information.
\newblock \emph{arXiv preprint arXiv:1703.00810}.

\bibitem[{Slobodkin et~al.(2023)Slobodkin, Goldman, Caciularu, Dagan, and Ravfogel}]{slobodkin2023curious}
Aviv Slobodkin, Omer Goldman, Avi Caciularu, Ido Dagan, and Shauli Ravfogel. 2023.
\newblock The curious case of hallucinatory unanswerablity: Finding truths in the hidden states of over-confident large language models.
\newblock \emph{arXiv preprint arXiv:2310.11877}.

\bibitem[{Socher et~al.(2013)Socher, Perelygin, Wu, Chuang, Manning, Ng, and Potts}]{socher2013recursive}
Richard Socher, Alex Perelygin, Jean Wu, Jason Chuang, Christopher~D Manning, Andrew~Y Ng, and Christopher Potts. 2013.
\newblock Recursive deep models for semantic compositionality over a sentiment treebank.
\newblock In \emph{Proceedings of the 2013 conference on empirical methods in natural language processing}, pages 1631--1642.

\bibitem[{Solove(2005)}]{solove2005taxonomy}
Daniel~J Solove. 2005.
\newblock A taxonomy of privacy.
\newblock \emph{U. Pa. l. Rev.}, 154:477.

\bibitem[{Sun et~al.(2024)Sun, Huang, Wang, Wu, Zhang, Gao, Huang, Lyu, Zhang, Li et~al.}]{sun2024trustllm}
Lichao Sun, Yue Huang, Haoran Wang, Siyuan Wu, Qihui Zhang, Chujie Gao, Yixin Huang, Wenhan Lyu, Yixuan Zhang, Xiner Li, et~al. 2024.
\newblock Trustllm: Trustworthiness in large language models.
\newblock \emph{arXiv preprint arXiv:2401.05561}.

\bibitem[{Sun et~al.(2023)Sun, Shen, Zhou, Zhang, Chen, Cox, Yang, and Gan}]{sun2023principle}
Zhiqing Sun, Yikang Shen, Qinhong Zhou, Hongxin Zhang, Zhenfang Chen, David Cox, Yiming Yang, and Chuang Gan. 2023.
\newblock Principle-driven self-alignment of language models from scratch with minimal human supervision.
\newblock \emph{arXiv preprint arXiv:2305.03047}.

\bibitem[{Tabassi(2023)}]{NIST}
Elham Tabassi. 2023.
\newblock \href {https://doi.org/https://doi.org/10.6028/NIST.AI.100-1} {Artificial intelligence risk management framework (ai rmf 1.0)}.

\bibitem[{Tenney et~al.(2019)Tenney, Xia, Chen, Wang, Poliak, McCoy, Kim, Van~Durme, Bowman, Das et~al.}]{tenney2019you}
Ian Tenney, Patrick Xia, Berlin Chen, Alex Wang, Adam Poliak, R~Thomas McCoy, Najoung Kim, Benjamin Van~Durme, Samuel~R Bowman, Dipanjan Das, et~al. 2019.
\newblock What do you learn from context? probing for sentence structure in contextualized word representations.
\newblock \emph{arXiv preprint arXiv:1905.06316}.

\bibitem[{Tian et~al.(2023)Tian, Wang, Zhang, Chen, and Du}]{tian2023joma}
Yuandong Tian, Yiping Wang, Zhenyu Zhang, Beidi Chen, and Simon Du. 2023.
\newblock Joma: Demystifying multilayer transformers via joint dynamics of mlp and attention.
\newblock \emph{arXiv preprint arXiv:2310.00535}.

\bibitem[{Tishby and Zaslavsky(2015)}]{tishby2015deep}
Naftali Tishby and Noga Zaslavsky. 2015.
\newblock Deep learning and the information bottleneck principle.
\newblock In \emph{2015 ieee information theory workshop (itw)}, pages 1--5. IEEE.

\bibitem[{Touvron et~al.(2023{\natexlab{a}})Touvron, Lavril, Izacard, Martinet, Lachaux, Lacroix, Rozi{\`e}re, Goyal, Hambro, Azhar et~al.}]{touvron2023llama1}
Hugo Touvron, Thibaut Lavril, Gautier Izacard, Xavier Martinet, Marie-Anne Lachaux, Timoth{\'e}e Lacroix, Baptiste Rozi{\`e}re, Naman Goyal, Eric Hambro, Faisal Azhar, et~al. 2023{\natexlab{a}}.
\newblock Llama: Open and efficient foundation language models.
\newblock \emph{arXiv preprint arXiv:2302.13971}.

\bibitem[{Touvron et~al.(2023{\natexlab{b}})Touvron, Martin, Stone, Albert, Almahairi, Babaei, Bashlykov, Batra, Bhargava, Bhosale et~al.}]{touvron2023llama}
Hugo Touvron, Louis Martin, Kevin Stone, Peter Albert, Amjad Almahairi, Yasmine Babaei, Nikolay Bashlykov, Soumya Batra, Prajjwal Bhargava, Shruti Bhosale, et~al. 2023{\natexlab{b}}.
\newblock Llama 2: Open foundation and fine-tuned chat models.
\newblock \emph{arXiv preprint arXiv:2307.09288}.

\bibitem[{Turner et~al.(2023)Turner, Thiergart, Udell, Leech, Mini, and MacDiarmid}]{turner2023activation}
Alex Turner, Lisa Thiergart, David Udell, Gavin Leech, Ulisse Mini, and Monte MacDiarmid. 2023.
\newblock Activation addition: Steering language models without optimization.
\newblock \emph{arXiv preprint arXiv:2308.10248}.

\bibitem[{Van~Aken et~al.(2019)Van~Aken, Winter, L{\"o}ser, and Gers}]{van2019does}
Betty Van~Aken, Benjamin Winter, Alexander L{\"o}ser, and Felix~A Gers. 2019.
\newblock How does bert answer questions? a layer-wise analysis of transformer representations.
\newblock In \emph{Proceedings of the 28th ACM international conference on information and knowledge management}, pages 1823--1832.

\bibitem[{Wang et~al.(2018)Wang, Singh, Michael, Hill, Levy, and Bowman}]{wang2018glue}
Alex Wang, Amanpreet Singh, Julian Michael, Felix Hill, Omer Levy, and Samuel~R Bowman. 2018.
\newblock Glue: A multi-task benchmark and analysis platform for natural language understanding.
\newblock \emph{arXiv preprint arXiv:1804.07461}.

\bibitem[{Wang et~al.(2023{\natexlab{a}})Wang, Chen, Pei, Xie, Kang, Zhang, Xu, Xiong, Dutta, Schaeffer et~al.}]{wang2024decodingtrust}
Boxin Wang, Weixin Chen, Hengzhi Pei, Chulin Xie, Mintong Kang, Chenhui Zhang, Chejian Xu, Zidi Xiong, Ritik Dutta, Rylan Schaeffer, et~al. 2023{\natexlab{a}}.
\newblock Decodingtrust: A comprehensive assessment of trustworthiness in gpt models.
\newblock In \emph{Thirty-seventh Conference on Neural Information Processing Systems Datasets and Benchmarks Track}.

\bibitem[{Wang et~al.(2021)Wang, Xu, Wang, Gan, Cheng, Gao, Awadallah, and Li}]{wang2021adversarial}
Boxin Wang, Chejian Xu, Shuohang Wang, Zhe Gan, Yu~Cheng, Jianfeng Gao, Ahmed~Hassan Awadallah, and Bo~Li. 2021.
\newblock Adversarial glue: A multi-task benchmark for robustness evaluation of language models.
\newblock In \emph{Advances in Neural Information Processing Systems}.

\bibitem[{Wang and Shu(2023)}]{wang2023backdoor}
Haoran Wang and Kai Shu. 2023.
\newblock Backdoor activation attack: Attack large language models using activation steering for safety-alignment.
\newblock \emph{arXiv preprint arXiv:2311.09433}.

\bibitem[{Wang et~al.(2023{\natexlab{b}})Wang, Hu, Hou, Chen, Zheng, Wang, Yang, Huang, Ye, Geng et~al.}]{wang2023robustness}
Jindong Wang, Xixu Hu, Wenxin Hou, Hao Chen, Runkai Zheng, Yidong Wang, Linyi Yang, Haojun Huang, Wei Ye, Xiubo Geng, et~al. 2023{\natexlab{b}}.
\newblock On the robustness of chatgpt: An adversarial and out-of-distribution perspective.
\newblock \emph{arXiv preprint arXiv:2302.12095}.

\bibitem[{Wang et~al.(2024)Wang, Zhang, Li, Tan, Wang, Ren, Jiang, and Qiu}]{wang2024inferaligner}
Pengyu Wang, Dong Zhang, Linyang Li, Chenkun Tan, Xinghao Wang, Ke~Ren, Botian Jiang, and Xipeng Qiu. 2024.
\newblock Inferaligner: Inference-time alignment for harmlessness through cross-model guidance.
\newblock \emph{arXiv preprint arXiv:2401.11206}.

\bibitem[{Wang et~al.(2022)Wang, Kordi, Mishra, Liu, Smith, Khashabi, and Hajishirzi}]{wang2022self}
Yizhong Wang, Yeganeh Kordi, Swaroop Mishra, Alisa Liu, Noah~A Smith, Daniel Khashabi, and Hannaneh Hajishirzi. 2022.
\newblock Self-instruct: Aligning language model with self generated instructions.
\newblock \emph{arXiv preprint arXiv:2212.10560}.

\bibitem[{Xu et~al.(2021)Xu, Liu, Li, Jain, and Tang}]{robustness_fairness_1}
Han Xu, Xiaorui Liu, Yaxin Li, Anil Jain, and Jiliang Tang. 2021.
\newblock To be robust or to be fair: Towards fairness in adversarial training.
\newblock In \emph{International conference on machine learning}, pages 11492--11501.

\bibitem[{Xu et~al.(2019)Xu, Zhang, Luo, Xiao, and Ma}]{xu2019frequency}
Zhi-Qin~John Xu, Yaoyu Zhang, Tao Luo, Yanyang Xiao, and Zheng Ma. 2019.
\newblock Frequency principle: Fourier analysis sheds light on deep neural networks.
\newblock \emph{arXiv preprint arXiv:1901.06523}.

\bibitem[{Yuan et~al.(2024)Yuan, Pang, Cho, Sukhbaatar, Xu, and Weston}]{yuan2024self}
Weizhe Yuan, Richard~Yuanzhe Pang, Kyunghyun Cho, Sainbayar Sukhbaatar, Jing Xu, and Jason Weston. 2024.
\newblock Self-rewarding language models.
\newblock \emph{arXiv preprint arXiv:2401.10020}.

\bibitem[{Zhang et~al.(2022)Zhang, Zhou, Wan, Zheng, Chang, and Hsieh}]{zhang2022improving}
Cenyuan Zhang, Xiang Zhou, Yixin Wan, Xiaoqing Zheng, Kai-Wei Chang, and Cho-Jui Hsieh. 2022.
\newblock Improving the adversarial robustness of nlp models by information bottleneck.
\newblock \emph{arXiv preprint arXiv:2206.05511}.

\bibitem[{Zhang et~al.(2024{\natexlab{a}})Zhang, Zhang, Li, Gao, Wang, Lu, Zhao, Qiao, and Shao}]{zhang2024psysafe}
Zaibin Zhang, Yongting Zhang, Lijun Li, Hongzhi Gao, Lijun Wang, Huchuan Lu, Feng Zhao, Yu~Qiao, and Jing Shao. 2024{\natexlab{a}}.
\newblock Psysafe: A comprehensive framework for psychological-based attack, defense, and evaluation of multi-agent system safety.
\newblock \emph{arXiv preprint arXiv:2401.11880}.

\bibitem[{Zhang et~al.(2024{\natexlab{b}})Zhang, Yao, Liang, and Xu}]{zhang2024random}
Zeliang Zhang, Wei Yao, Susan Liang, and Chenliang Xu. 2024{\natexlab{b}}.
\newblock Random smooth-based certified defense against text adversarial attack.
\newblock In \emph{Findings of the Association for Computational Linguistics: EACL 2024}, pages 1251--1265.

\bibitem[{Zhao et~al.(2023)Zhao, Chen, Yang, Liu, Deng, Cai, Wang, Yin, and Du}]{zhao2023explainability}
Haiyan Zhao, Hanjie Chen, Fan Yang, Ninghao Liu, Huiqi Deng, Hengyi Cai, Shuaiqiang Wang, Dawei Yin, and Mengnan Du. 2023.
\newblock Explainability for large language models: A survey.
\newblock \emph{ACM Transactions on Intelligent Systems and Technology}.

\bibitem[{Zhou et~al.(2023{\natexlab{a}})Zhou, Liu, Xu, Iyer, Sun, Mao, Ma, Efrat, Yu, Yu et~al.}]{zhou2023lima}
Chunting Zhou, Pengfei Liu, Puxin Xu, Srini Iyer, Jiao Sun, Yuning Mao, Xuezhe Ma, Avia Efrat, Ping Yu, Lili Yu, et~al. 2023{\natexlab{a}}.
\newblock Lima: Less is more for alignment.
\newblock \emph{arXiv preprint arXiv:2305.11206}.

\bibitem[{Zhou et~al.(2024)Zhou, Zhang, Deng, Liu, Shen, Chan, and Zhang}]{zhou2024explaining}
Huilin Zhou, Hao Zhang, Huiqi Deng, Dongrui Liu, Wen Shen, Shih-Han Chan, and Quanshi Zhang. 2024.
\newblock Explaining generalization power of a dnn using interactive concepts.
\newblock In \emph{Proceedings of the AAAI Conference on Artificial Intelligence}, volume~38, pages 17105--17113.

\bibitem[{Zhou and Srikumar(2022)}]{zhou2022closer}
Yichu Zhou and Vivek Srikumar. 2022.
\newblock A closer look at how fine-tuning changes bert.
\newblock In \emph{Proceedings of the 60th Annual Meeting of the Association for Computational Linguistics (Volume 1: Long Papers)}, pages 1046--1061.

\bibitem[{Zhou et~al.(2023{\natexlab{b}})Zhou, Zhou, Li, Yao, Yao, and Han}]{zhou2023mcgra}
Zhanke Zhou, Chenyu Zhou, Xuan Li, Jiangchao Yao, Quanming Yao, and Bo~Han. 2023{\natexlab{b}}.
\newblock On strengthening and defending graph reconstruction attack with markov chain approximation.
\newblock In \emph{International Conference on Machine Learning}.

\bibitem[{Zhu et~al.(2023)Zhu, Wang, Zhou, Wang, Chen, Wang, Yang, Ye, Gong, Zhang et~al.}]{zhu2023promptbench}
Kaijie Zhu, Jindong Wang, Jiaheng Zhou, Zichen Wang, Hao Chen, Yidong Wang, Linyi Yang, Wei Ye, Neil~Zhenqiang Gong, Yue Zhang, et~al. 2023.
\newblock Promptbench: Towards evaluating the robustness of large language models on adversarial prompts.
\newblock \emph{arXiv preprint arXiv:2306.04528}.

\bibitem[{Zou et~al.(2023)Zou, Phan, Chen, Campbell, Guo, Ren, Pan, Yin, Mazeika, Dombrowski et~al.}]{zou2023representation}
Andy Zou, Long Phan, Sarah Chen, James Campbell, Phillip Guo, Richard Ren, Alexander Pan, Xuwang Yin, Mantas Mazeika, Ann-Kathrin Dombrowski, et~al. 2023.
\newblock Representation engineering: A top-down approach to ai transparency.
\newblock \emph{arXiv preprint arXiv:2310.01405}.

\end{thebibliography}

\newpage
\appendix
\onecolumn

\tableofcontents

\newpage

{\centering \section*{Appendix}}

\section{Guidelines for Trustworthy LLMs}
\label{appendix:guideline}
The surge of LLMs brings significant concerns regarding their trustworthiness, especially considering the security risks inherent in the models themselves and the agents based on these models~\cite{wang2023backdoor,ji2023beavertails,CLTC,NIST,li2024salad,zhang2024psysafe}, which pertains to the aspects and extent to which humans can trust AI. 
% In understanding and achieving trustworthiness, 
Existing research in AI governance and trustworthy LLMs provides guidance for establishing comprehensive and reliable dimensions of trustworthy LLMs in this study.

% AI governance
Governments~\cite{NIST,doi/10.2759/346720}, organizations~\cite{AI_Act,AI_Verify}, and research institutions~\cite{CLTC,liu2023trustworthy} worldwide have proposed classifications from various perspectives such as the AI lifecycle, the acceptability of AI risk, considering AI governance at different levels including individual, institutional, and societal. Among these, categories stemming from the technological aspect offer guidance for trustworthy AI~\cite{Liu2023trustworthy_ai}, such as robustness, fairness, accountability, transparency, etc.

% Trustworthy LLM

% category
By integrating AI governance principles into trustworthy LLMs, not only aids in developing more credible LLMs but also promotes the sustainable and responsible application of AI technology. Concurrently, taking into account the categorizations of trustworthy LLMs~\cite{liu2023trustworthy,wang2024decodingtrust} and prioritizing both adherence to principles and addressing practical challenges faced by LLMs, six primary categories have been identified: robustness, reliability, fairness, toxicity, privacy, and interpretability. In this study, interpretability is employed as a tool to explore the other five concepts of trustworthiness.

\section{Datasets of Truthworthy LLMs}
\label{appendix:dataset}
Considering five aspects of trustworthiness: reliability, toxicity, privacy, fairness, and robustness, we carefully design five binary NLP datasets. These datasets are tailored from independent lines of trustworthy AI research, with labels indicating whether a sentence satisfies each aforementioned aspect of trustworthiness. In other words, the label indicates whether the corresponding sentence contains untrue (or unfair, toxic, privacy-leakage, and perturbed) information. 

The datasets considered below are balanced, i.e., the number of positive and negative numbers are almost the same. In other words, some special cases, for example, the random classifier on these datasets, will achieve an accuracy of around 50\%.

\noindent
\paragraph{Reliability.} We use TruthfulQA~\cite{Lin_Hilton_Evans_2022} to measure the truthfulness modeling ability of LLMs. TruthfulQA comprises 817 questions across 38 categories, designed to evaluate the veracity of answers generated by language models. We concatenate the multiple-choice questions and their respective candidate answers to form either correct or incorrect statements, which is used to measure the reliability of large language models in discerning truthfulness.

\noindent
\paragraph{Toxicity.} We choose ToxiGen~\cite{hartvigsen2022toxigen} to measure the toxicity modeling ability of LLMs. ToxiGen is a large-scale dataset encompassing a range of implicit toxic and non-toxic statements associated with 13 minority demographics. 
Following Llama2~\cite{touvron2023llama},  we employ a revised version of the dataset from~\cite{hosseini-etal-2023-empirical}, selectively retaining those sentences that achieved unanimous agreement from the annotators regarding the target demographic group.

\noindent
\paragraph{Privacy.} We choose the tier 2 task from ConfAIde~\cite{mireshghallah2023llms} to measure the privacy awareness of LLMs. ConfAIde focuses on contextual privacy and aims to pinpoint key vulnerabilities in LLMs' privacy reasoning abilities. Given the limited data volume, we constructed new data based on ConfAIde and the Solove Taxonomy~\cite{solove2005taxonomy} to assess the privacy awareness of LLMs regarding given information. Solove Taxonomy comprises 4 major categories and 16 subcategories. For each subcategory, we designed prompts and provided 2 to 6 examples to facilitate data generation using GPT-4. The generated data were then assessed by GPT-4 for privacy violations, selecting entries with high confidence (consistent judgments in five assessments). We combined generated data with ConfAIde to consider whether LLMs can identify privacy violations.

\noindent
\paragraph{Fairness.} We use StereoSet~\citep{stereoset} to measure the stereotype modeling ability of LLMs, i.e., whether LLMs capture stereotypical biases about race, religion, profession, and gender. Taking inter-sentence tests as the original dataset, we concatenate the context and the candidate sentence into one sentence, and the corresponding class label follows the candidate sentences, capturing stereotypical, anti-stereotypical, and unrelated associations. We assign a binary label to every sentence to indicate whether it contains stereotypical bias.

\noindent
\paragraph{Robustness.} Following the construction of AdvGLUE benchmark~\citep{wang2021adversarial}, we perturb GLUE benchmark~\citep{wang2018glue} in a human-imperceptible way. Specifically, we select SST-2~\citep{socher2013recursive} from GLUE. It is a popular dataset in robustness literature~\citep{zhu2023promptbench,zhang2022improving,zhang2024random,chen2024taichi}.
We introduce typos by randomly changing the case of 20\% letters in each sentence from the SST-2~\citep{socher2013recursive} validation set. We assign a binary label to every sentence to indicate whether it has been attacked.

% \clearpage
\section{More Detailed Experimental Settings}
\label{appendix:exp-setting}

% setting
\begin{table*}[t!]
\centering
\caption{Summary of experimental settings related to trustworthiness datasets.}
\vspace{1mm}
\label{table:appen-settings}
\setlength{\tabcolsep}{1mm}
\scalebox{0.87}{
    \begin{tabular}{p{2cm}p{3cm}p{3cm}p{3cm}p{3cm}p{3cm}}
    \toprule
    Dimension & Reliability & Toxicity & Privacy & Fairness  & Robustness \\ 
    \toprule
    Benchmark & TruthfulQA & ToxiGen & ConfAIde & StereoSet  & SST-2 \\ 
    \midrule
    Evaluation Metrics & Truth\% and Info\% & Toxic Ratio & Accuracy & Accuracy  & Accuracy \\
    \midrule
    The meaning of labels in activation datasets & $y=0$: statements with false answer\newline
    $y=1$: statements with true answer 
    & $y=0$: toxic statements\newline
    $y=1$: benign statements 
    & $y=0$: statements that do not conclude privacy violation\newline
    $y=1$: statements that conclude privacy violation & $y=0$: benign statements\newline
    $y=1$: stereotypical statements  
    & $y=0$: the original sentence\newline
    $y=1$: the perturbed sentence\\
    \bottomrule
    \end{tabular}
}
\end{table*}

\paragraph{Dataset partition.}
Within each dataset, following~\cite{li2023inferencetime}, we first split the original dataset into a development set and a test set at a 1:1 ratio. We further divide the development set into a training/validation set at a 4:1 ratio for the training and evaluation of the linear probe, with the steering vector also being constructed based on the development set. The test set is used to assess model performance, ensuring no data leakage occurs during the experiment.

\paragraph{Evaluation on trustworthiness abilities benchmarks.}
For TruthfulQA, we adopt the QA prompts following InstructGPT ~\cite{ouyang2022training}. Additionally, two fine-tuned GPT-3 models, i.e. a ``GPT-judge''\footnote{ft:davinci-002:zy-pj-035:truthfulqa-truth:8nKPYSTt} and a ``GPT-info,''\footnote{ft:davinci-002:zy-pj-035:truthfulqa-info:8nJbtN57} are used to predict the truthfulness and informativeness of the generated outputs from LLMs, respectively.
For ToxiGen, we follow~\cite{touvron2023llama}, employing the default ToxiGen classifier~\cite{hartvigsen2022toxigen} fine-tuned on RoBERTa~\cite{liu2020roberta} to evaluate the toxicity of contents generated by LLMs, and finally reporting the proportion of generated text classified as toxic.
% \highlight{ConfAIde @zj}
For ConfAIde, we use the tier 2 task to assess the agreement on privacy information usage. We employ the same evaluation prompt as ConfAIde~\cite{mireshghallah2023llms}, with the adaptation of converting multiple-choice questions into binary classification tasks to evaluate the accuracy.
For StereoSet, following TrustLLM~\citep{sun2024trustllm}, 
we provide prompts using the same template for the stereotype recognition task as theirs. The generated choices are then compared with the ground-truth labels to obtain accuracy. For perturbed SST-2, we follow~\citet{wang2023robustness} and use the same prompt as theirs.
TruthfulQA is evaluated in a 6-shot setting, whereas other benchmarks are conducted in 0-shot settings.

\paragraph{Evaluation on general abilities benchmarks.}
For all the results on ARC, MMLU, MathQA, and RACE reported in Section~\ref{sec:steering} of the main body, we conduct evaluations using the lm-evaluation-harness library~\cite{eval-harness} with its default evaluation settings.

\paragraph{Selection of perplexity.}
Regarding perplexity, we follow~\cite{radford2019language} to calculate the perplexity on LAMBADA~\cite{paperno-etal-2016-lambada}. The perplexity value reported for GPT-2 in~\cite{radford2019language} is 8.6, and the perplexity we tested for AmberChat is 4.5. Based on our observations, we consider a perplexity value of less than 6 to be a reasonable threshold, please refer to Appendix~\ref{appendix-ppl-case} for examples.

\paragraph{Reproduce the first pre-training checkpoint.}
In our initial experimental observations using the pre-training checkpoints released in~\cite{liu2023llm360}, we noticed that the mutual information $I(T, X)$ appeared to be consistently decreasing, which contradicts the existing two-phase phenomenon~\cite{shwartz2017opening}. This led us to speculate the possibility of overlooked experimental insights between the initial model state and the first checkpoint. Therefore, to observe more finer-grained dynamics during the pre-training phase, we utilized the official code released by~\cite{liu2023llm360}\footnote{\url{https://github.com/LLM360/amber-train}}, ensuring the hyperparameters are consistent with those reported in the original paper. We initiated pre-training from a randomly initialized model using the corpus for the first checkpoint and saved more finely-grained checkpoints to observe finer experimental phenomena.

% \clearpage
\section{Full Linear Probing Results}
\label{appendix:full-probing}

\begin{figure*}[h]
    \centering
    \includegraphics[width=0.7\linewidth]{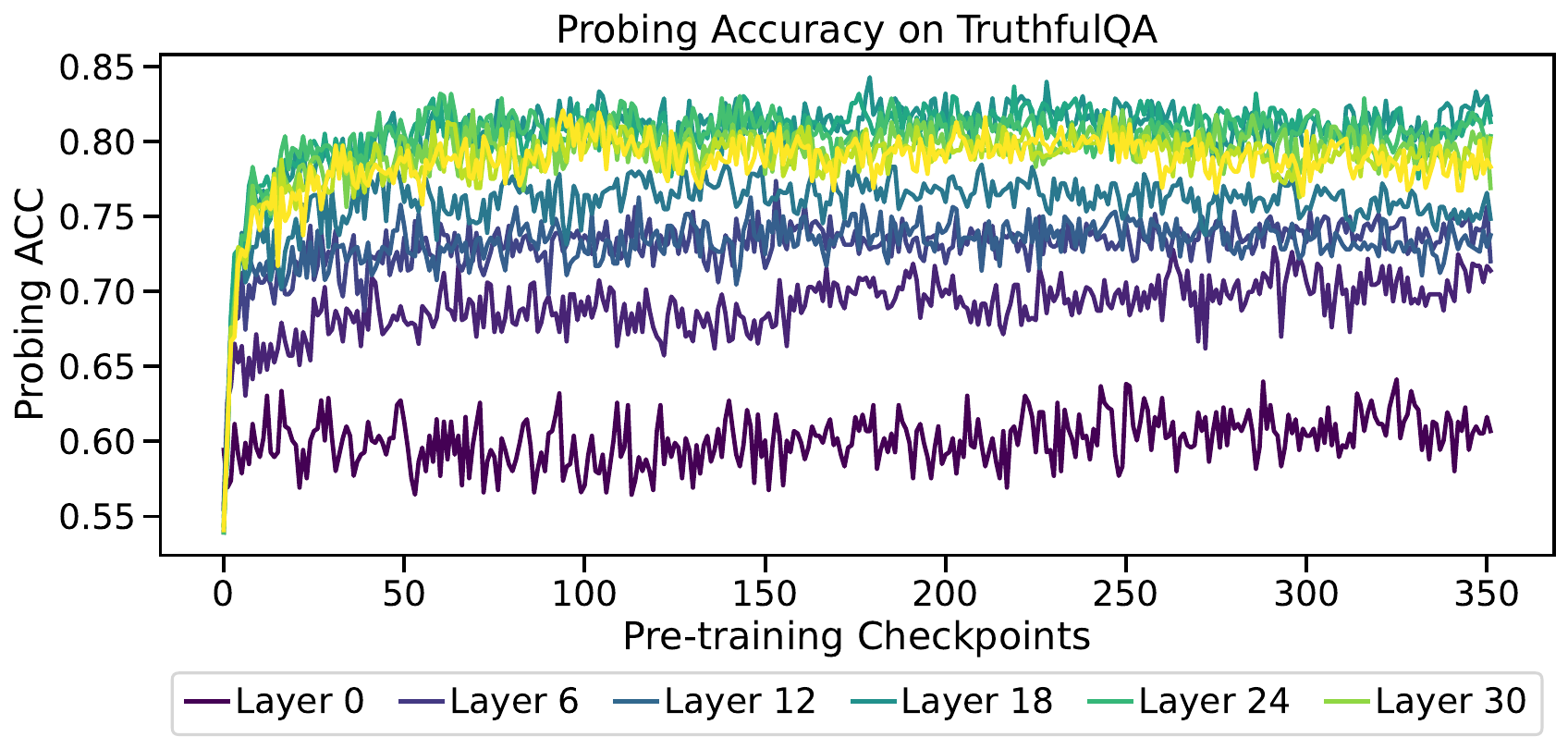}
    \caption{The linear probe accuracy on TruthfulQA for all 360 pre-training checkpoints.}
    \label{fig:appen-probing-full-tqa}
\end{figure*}

\begin{figure*}[h]
    \centering
    \includegraphics[width=0.7\linewidth]{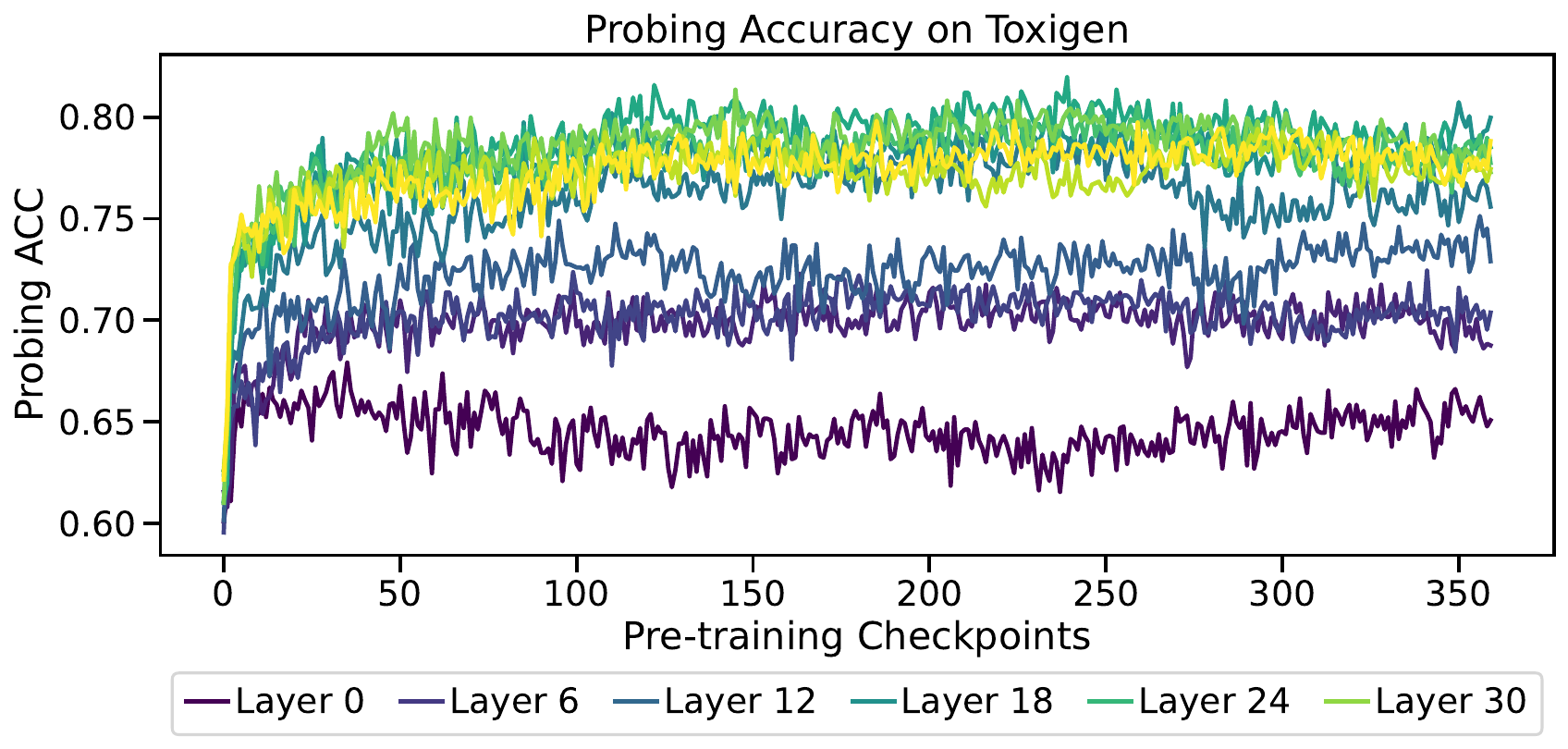}
    \caption{The linear probe accuracy on Toxigen for all 360 pre-training checkpoints.}
    \label{fig:appen-probing-full-toxigen}
\end{figure*}

\begin{figure*}[h]
    \centering
    \includegraphics[width=0.7\linewidth]{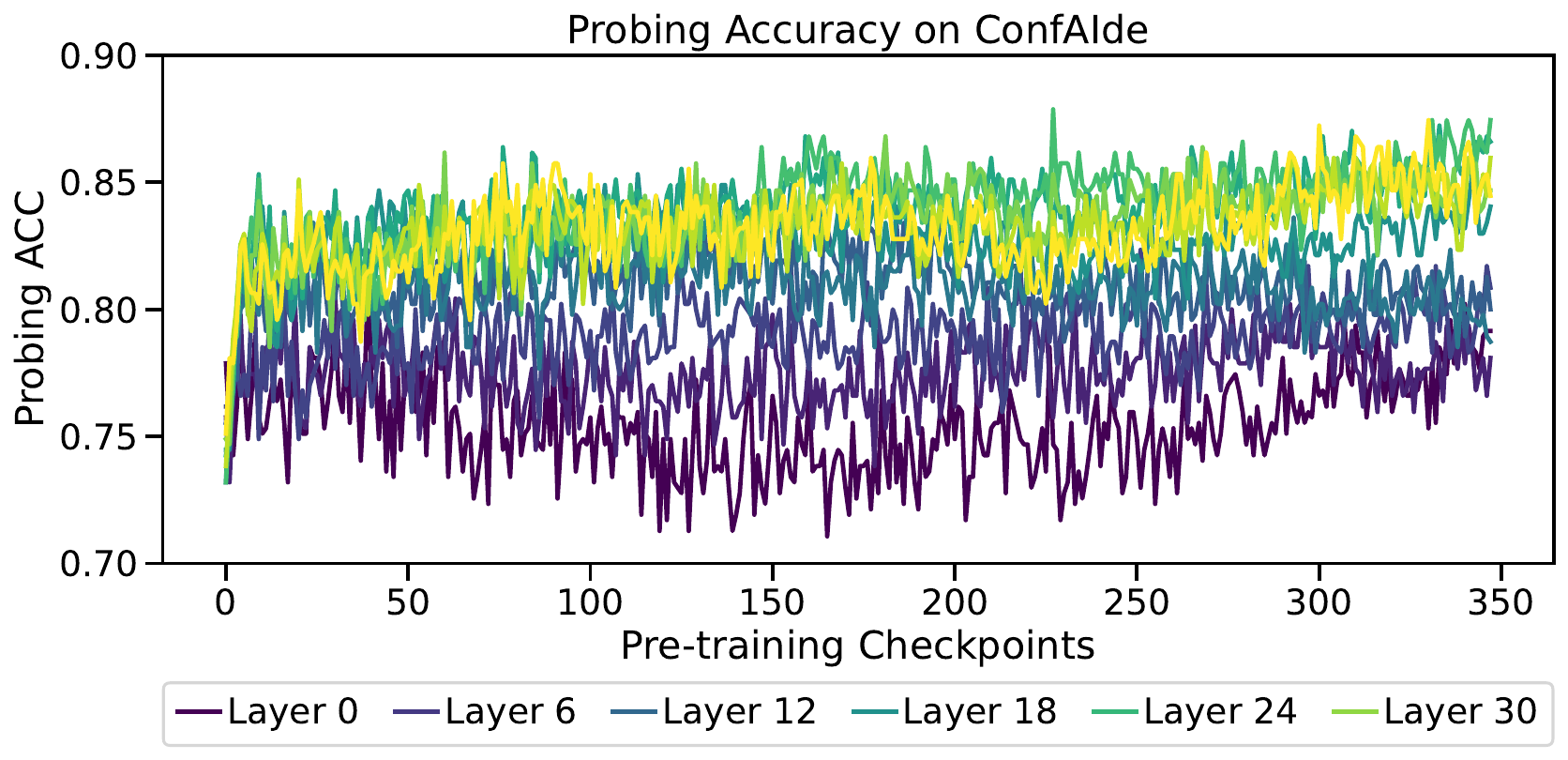}
    \caption{The linear probe accuracy on ConfAIde for all 360 pre-training checkpoints.}
    \label{fig:appen-probing-full-privacy}
\end{figure*}

\begin{figure*}[h]
    \centering
    \includegraphics[width=0.7\linewidth]{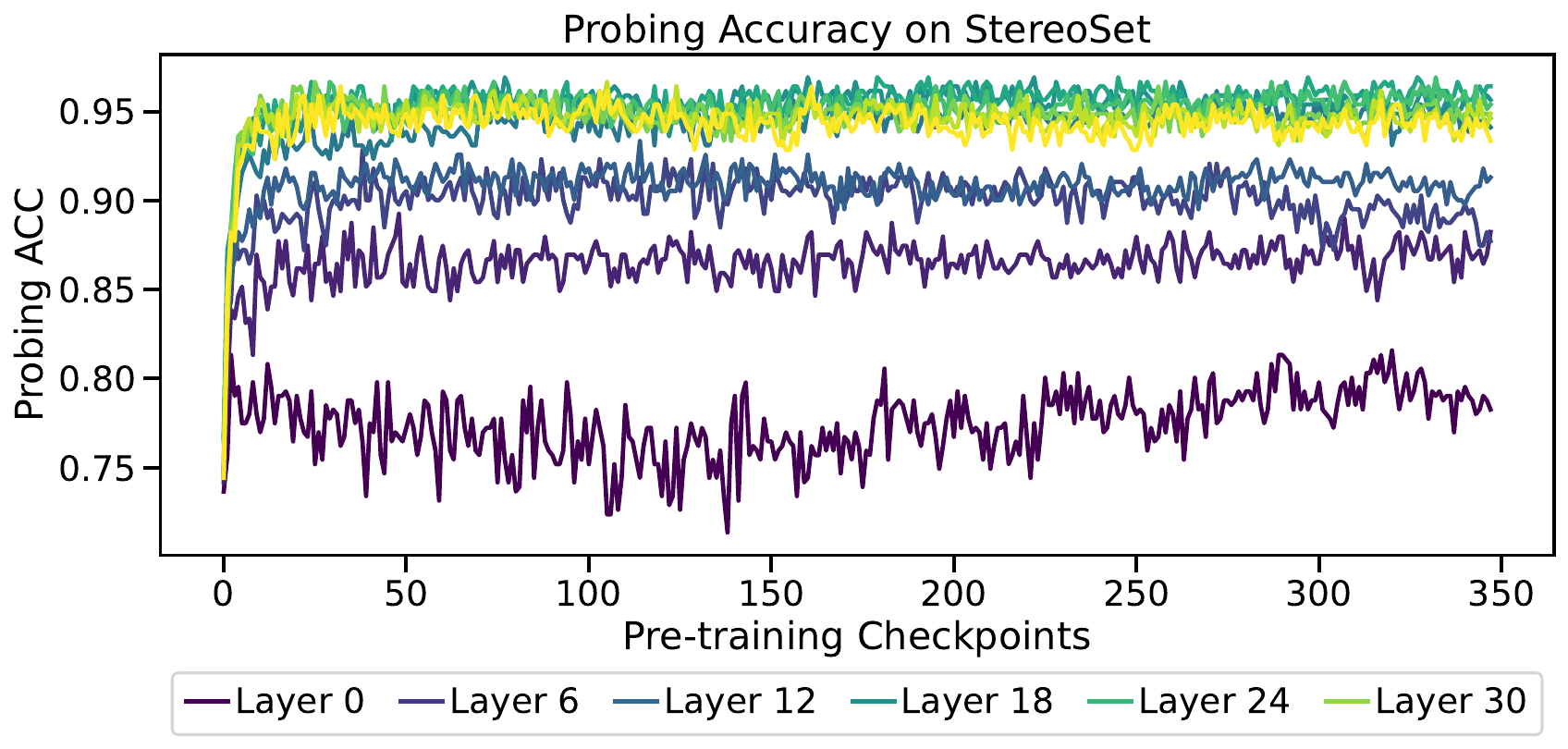}
    \caption{The linear probe accuracy on StereoSet for all 360 pre-training checkpoints.}
    \label{fig:appen-probing-full-fairness}
\end{figure*}

\begin{figure*}[h]
    \centering
    \includegraphics[width=0.7\linewidth]{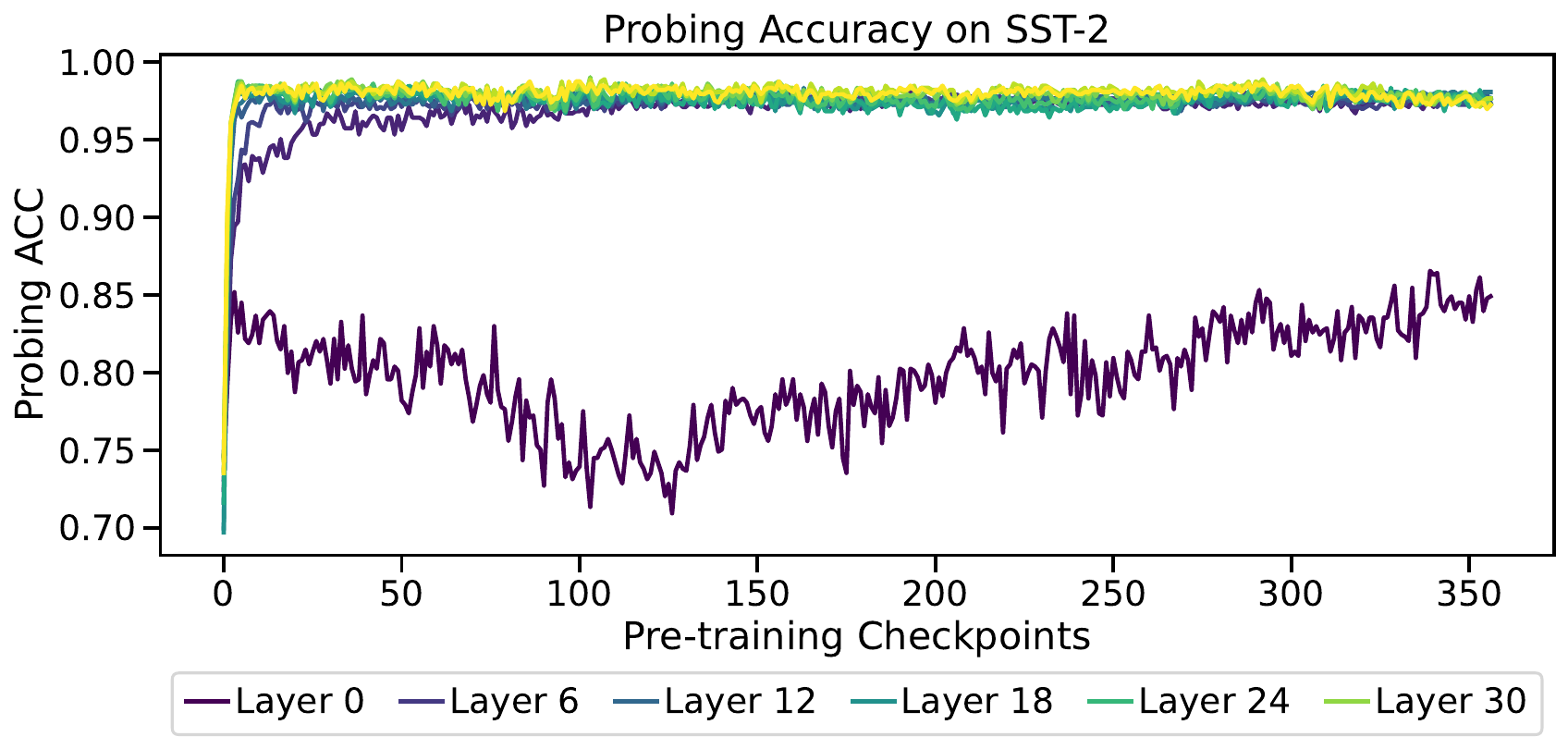}
    \caption{The linear probe accuracy on SST-2  for all 360 pre-training checkpoints.}
    \label{fig:appen-probing-full-robustness}
\end{figure*}

The full linear probing results from 360 checkpoints in five trustworthiness dimensions are shown in Figure~\ref{fig:appen-probing-full-tqa},\ref{fig:appen-probing-full-toxigen},\ref{fig:appen-probing-full-privacy},\ref{fig:appen-probing-full-fairness},\ref{fig:appen-probing-full-robustness}. 
Overall, the experimental observations and conclusions are consistent with Section~\ref{subsec:probing_results}.
Results from five datasets together suggest that middle-layer representations exhibit linearly separable patterns. 
Furthermore, the probing accuracy increases during the initial phase of pre-training, followed by fluctuation throughout the remaining pre-training period.

%% fig: probing-full-tqa

%% fig: probing-full-toxien

%% fig: probing-full-privacy

%% fig: probing-full-fairness

%% fig: probing-full-robustness

\clearpage

\section{Supplementary Details for `Probing LLM using Mutual Information'} \label{appendix:sec:IB_LLM}

\subsection{Mutual Information and HSIC}

\begin{definition}[Mutual Information (MI)]
Given two continuous random variables $X$ and $Y$, the mutual information is defined as:
\begin{align}
I(X ; Y)=\int_Y \int_X p(x, y) \log \frac{p(x, y)}{p(x) p(y)} d x d y.
\end{align}
\end{definition}
Mutual information is a measure of the mutual dependence between the two variables. However, because of the difficulty to accurately compute mutual information~\citep{kraskov2004estimating}, we follow~\citet{ma2020hsic} to use HSIC~\citep{gretton2005measuring} as an estimator of mutual information. HSIC~\citep{gretton2005measuring} also indicates the dependency between two random variables. For other kinds of estimation, please refer to Appendix E.3 in \citet{zhou2023mcgra}.

\begin{definition}[Hilbert-Schmidt Independence Criterion (HSIC)]
It is the Hilbert-Schmidt norm of the cross-covariance operator between the distributions in Reproducing Kernel Hilbert Space (RKHS). $\operatorname{HSIC}(X,Y)$ is defined as:
% \footnote{The definition of HSIC regarding the expectation may be omitted. The empirical form is enough.}
\begin{align}
\operatorname{HSIC}(X, Y) & =\mathbb{E}_{X Y X^{\prime} Y^{\prime}}\left[k_X\left(X, X^{\prime}\right) k_{Y^{\prime}}\left(Y, Y^{\prime}\right)\right] \notag \\
& +\mathbb{E}_{X X^{\prime}}\left[k_X\left(X, X^{\prime}\right)\right] \mathbb{E}_{Y Y^{\prime}}\left[k_Y\left(Y, Y^{\prime}\right)\right] \notag \\
& -2 \mathbb{E}_{X Y}\left[\mathbb{E}_{X^{\prime}}\left[k_X\left(X, X^{\prime}\right)\right] \mathbb{E}_{Y^{\prime}}\left[k_Y\left(Y, Y^{\prime}\right)\right]\right],
\end{align}
where $X^{\prime}$, $Y^{\prime}$ are independent copies of $X$, $Y$, respectively, and $k_X$ , $k_Y$ are kernels. 
\end{definition}

$\operatorname{HSIC}(X,Y)$ is zero if and only if the random variables $X$ and $Y$ are independent. In practice, given the activation dataset $\mathcal{D}$, we empirically estimate HSIC as
\begin{align}
    \widehat{\operatorname{HSIC}}(X, Y)=(n-1)^{-2} \operatorname{tr}\left(K_X H K_Y H\right),
\end{align}
where $K_X$ and $K_Y$ are kernel matrices with entries $K_{X_{i j}}=k_X\left(x_i, x_j\right)$ and $K_{Y_{i j}}=k_Y\left(y_i, y_j\right)$, respectively, and $H=\mathbf{I}-\frac{1}{n} \mathbf{1} \mathbf{1}^{\top}$ is a centering matrix. Following~\citep{ma2020hsic}, we choose Gaussian kernel $k(\mathbf{x}, \mathbf{y}) \sim \exp \left(-\frac{1}{2}\|\mathbf{x}-\mathbf{y}\|^2 / \sigma^2\right)$. The scaling parameter $\sigma$ is selected by grid search in $[50,400]$.

\subsection{Mutual Information Results across Five Trustworthiness Dimensions} \label{appendix:mi_others}

\begin{figure*}[b]
    \centering
    \includegraphics[width=0.8\linewidth]{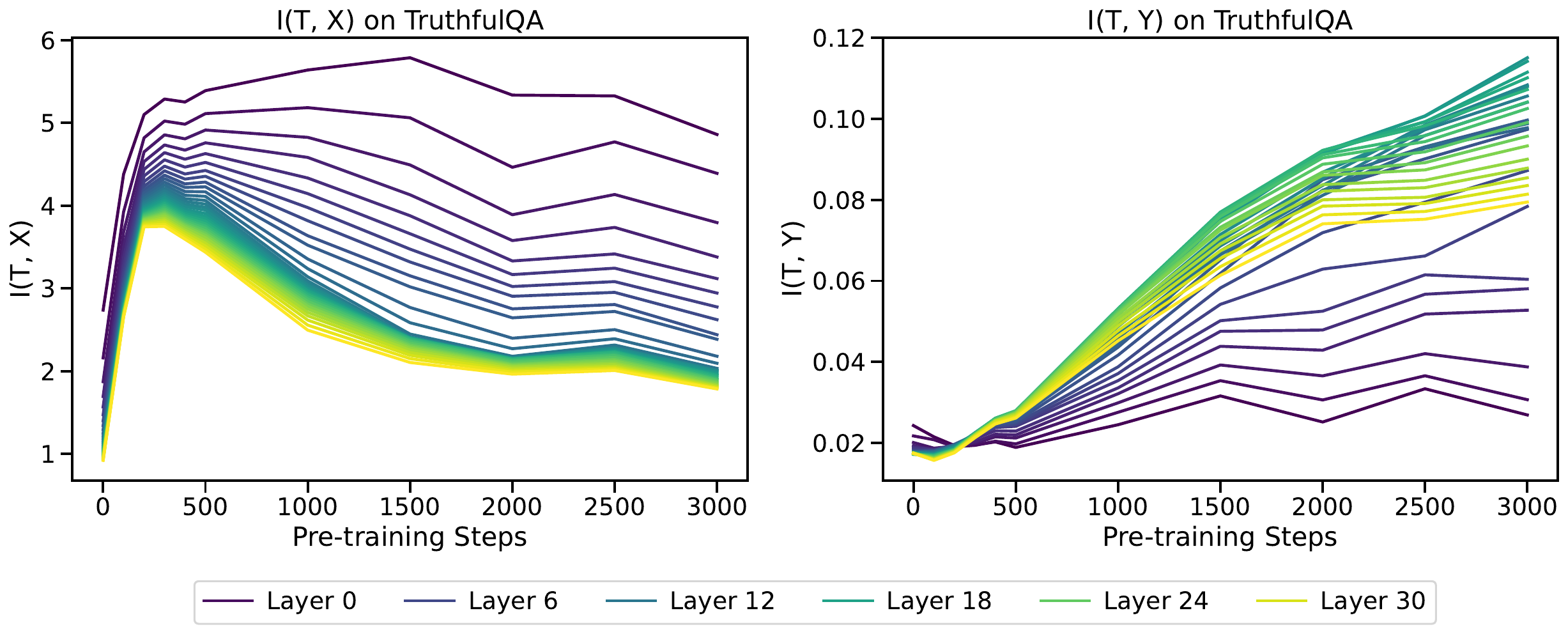}
    \caption{The dynamics of $I(T, X)$ and $I(T, Y)$ for TruthfulQA across various layers during pre-training.}
    \label{fig:appen-mi-full-tqa}
\end{figure*}

\begin{figure*}[t]
    \centering
    \includegraphics[width=0.8\linewidth]{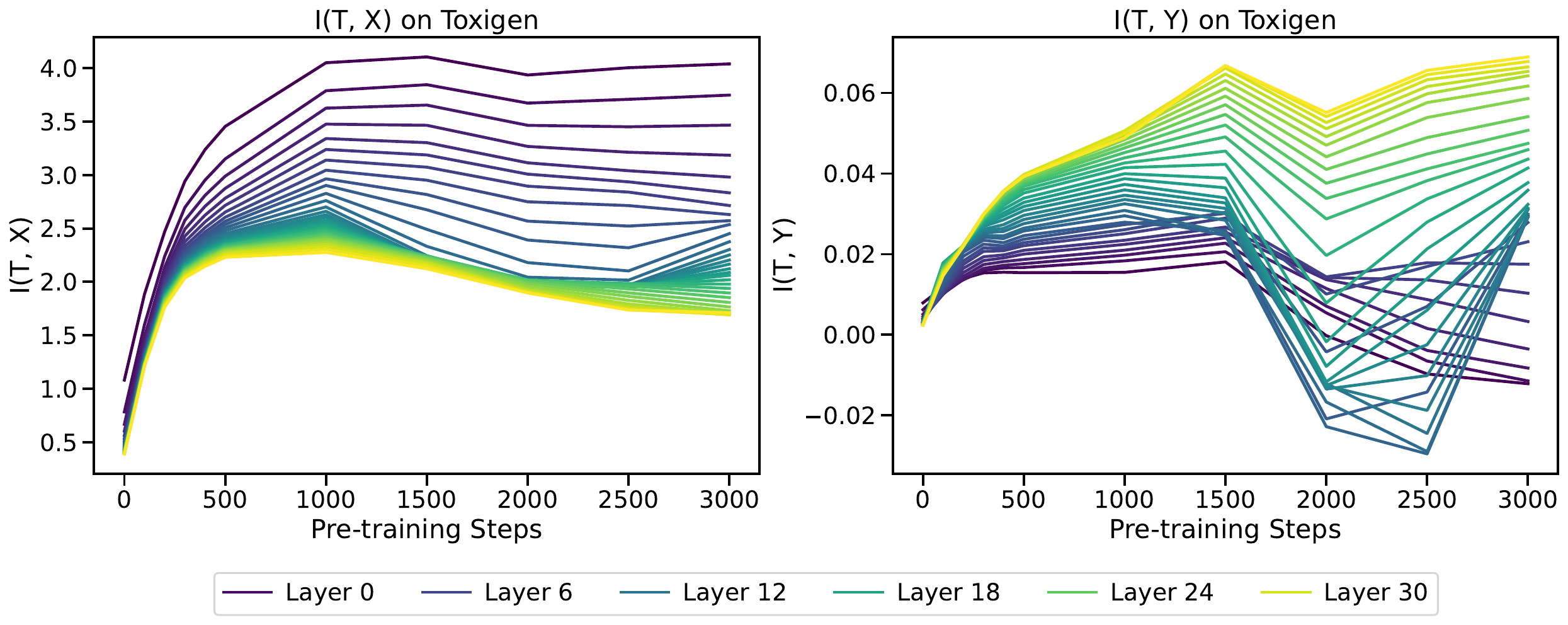}
    \caption{The dynamics of $I(T, X)$ and $I(T, Y)$ for Toxigen across various layers during pre-training.}
    \label{fig:appen-mi-full-toxigen}
\end{figure*}

\begin{figure*}[t]
    \centering
    \includegraphics[width=0.8\linewidth]{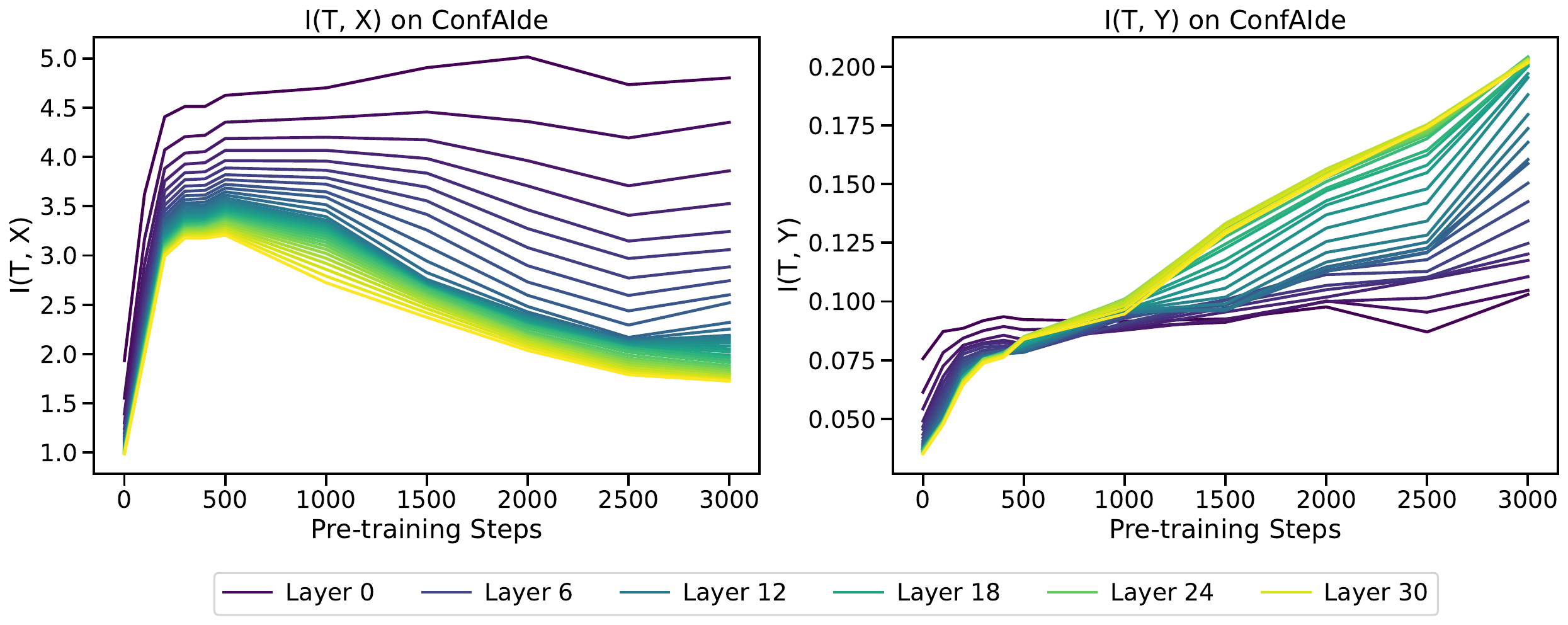}
    \caption{The dynamics of $I(T, X)$ and $I(T, Y)$ for ConfAIde across various layers during pre-training.}
    \label{fig:appen-mi-full-privacy}
\end{figure*}

\begin{figure*}[t]
    \centering
    \includegraphics[width=0.8\linewidth]{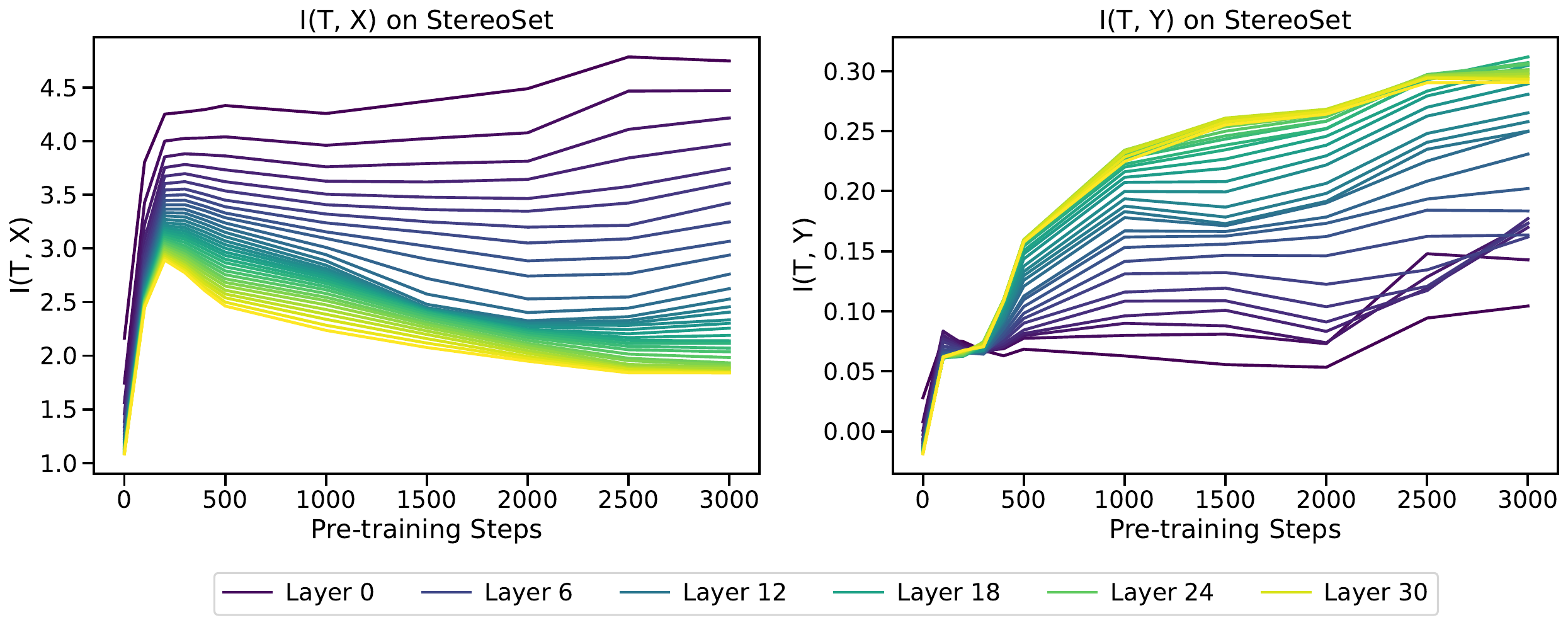}
    \caption{The dynamics of $I(T, X)$ and $I(T, Y)$ for StereoSet across various layers during pre-training.}
    \label{fig:appen-mi-full-fairness}
\end{figure*}

\begin{figure*}[h]
    \centering
    \includegraphics[width=0.8\linewidth]{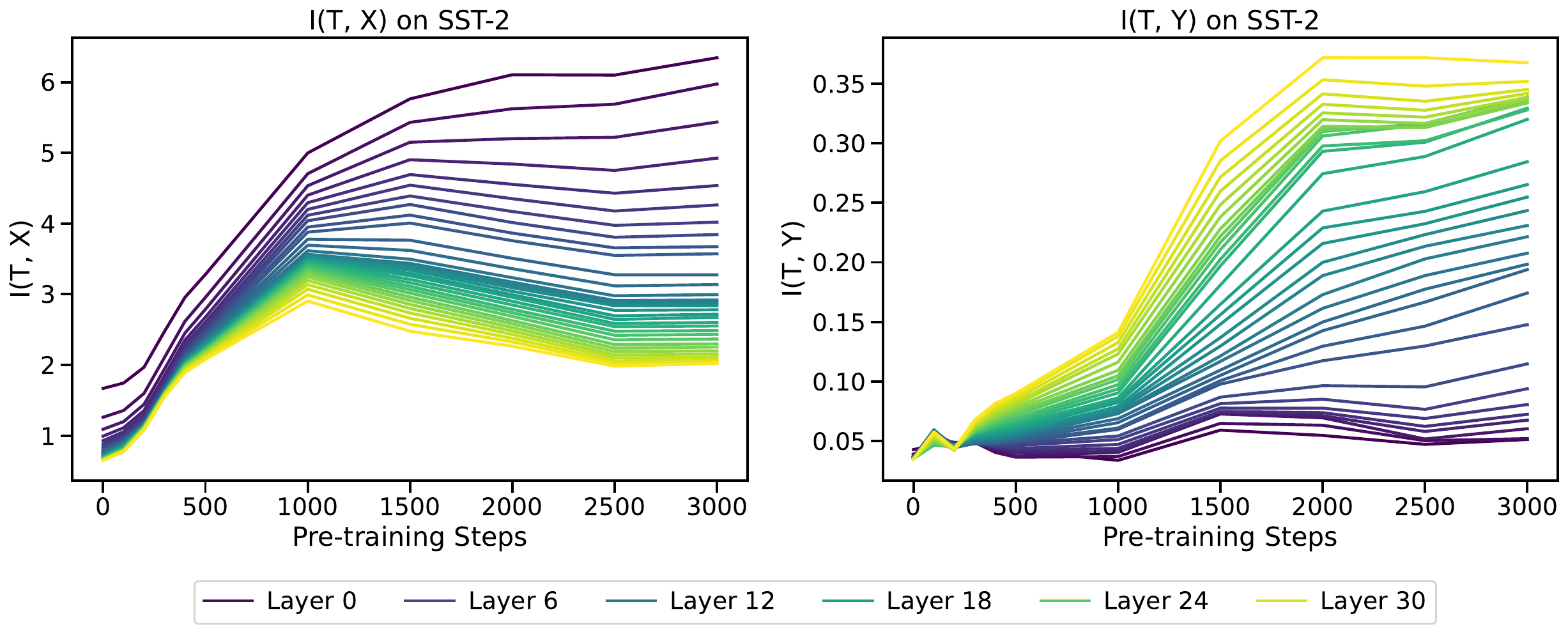}
    \caption{The dynamics of $I(T, X)$ and $I(T, Y)$ for SST-2 across various layers during pre-training.}
    \label{fig:appen-mi-full-robustness}
\end{figure*}

Figure~\ref{fig:appen-mi-full-tqa},\ref{fig:appen-mi-full-toxigen},\ref{fig:appen-mi-full-privacy},\ref{fig:appen-mi-full-fairness},\ref{fig:appen-mi-full-robustness} show the trend of mutual information on five trustworthiness dimensions.
% from all 360 pre-training checkpoints during the full pre-training period. 
The results are also consistent with the dynamics in Section~\ref{subsec:information_result}. The phase transition from ``fitting'' to ``compression'' is also applicable: there are also two phases during pre-training. In the first and shorter phase, both $I(T, X)$ and $I(T, Y)$ increase. While in the second and much longer phase, $I(T, X)$ decreases, and $I(T, Y)$ continues to increase. There are some fluctuations of $I(T, Y)$ for Toxigen, which may be due to the instability of pre-training.

% \clearpage
%% fig: mi: tqa

%% fig: mi: toxigen

%% fig: mi: privacy

%% fig: mi: fairness

\clearpage
%% fig: mi: robustness

\section{Experimental Results on Another Series of LLMs' Checkpoints}
\label{appendix-olmo}

To further demonstrate the generalization performance of the observations in this work, we conduct additional experiments on a recently released open-source model named OLMo~\cite{groeneveld2024olmo}. OLMo provides all intermediate checkpoints during the pre-training period. Additionally, OLMo also releases an instruction-tuned model named OLMo-7B-SFT and an aligned model through DPO named OLMo-7B-Instruct. Currently, the largest model size available in the OLMo project is 7B parameters.

\begin{figure*}[ht]
  \centering
\includegraphics[width=\linewidth]{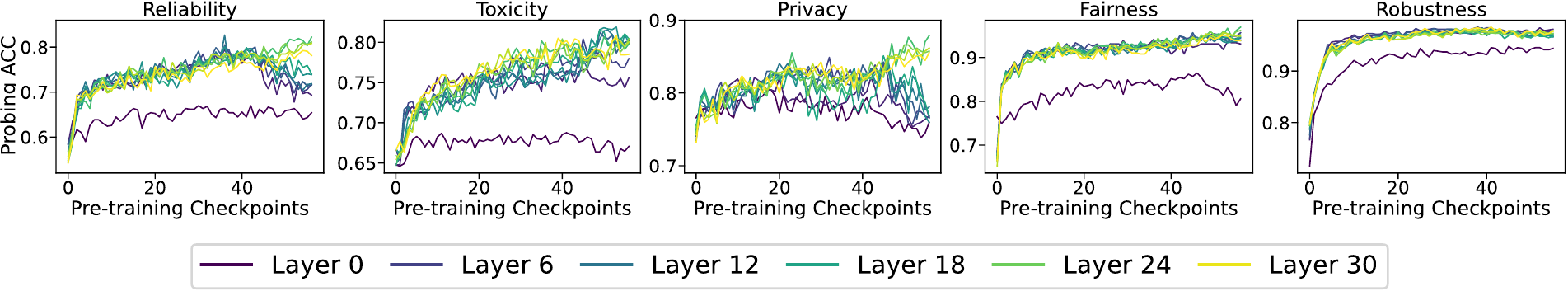}
% \vspace{-15pt}
\caption{The linear probe accuracy on five trustworthiness dimensions for the first 60 pre-training checkpoints of OLMo. For each checkpoint, we report the results from layers \{0, 6, 12, 18, 24, 30\}.
}
\label{fig:olmo-probing-all}
\end{figure*}

\noindent
We follow the experimental settings in Section~\ref{sec:methods} to conduct probing experiments on OLMo. The results are shown in Figure~\ref{fig:olmo-probing-all}.
Figure~\ref{fig:olmo-probing-all} demonstrates that after the early pre-training period, middle layer representations of LLMs have already developed linearly separable patterns about trustworthiness. This aligns with the results obtained from LLM360, as introduced in Section~\ref{subsec:probing_results}.

% steering vector - OLMo on truthfulqa and fairness 
\begin{table*}[h]
\centering
% \vspace{-3pt}
\caption{Results of activation intervention on OLMo in TruthfulQA and StereoSet.
Format and significance markers remain consistent with Table~\ref{table:tqa}. $\bm v_{ckpt\_{279}}$ denotes the steering vector extracted from the $279$-th checkpoint.
}
\vspace{1mm}
\label{table:steering-olmo}
\setlength{\tabcolsep}{2.5mm}
\scalebox{0.85}{
    \begin{tabular}{l|ccc|c}
    \toprule
    \multirow{2}{*}{Model}  & \multicolumn{3}{c|}{TruthfulQA Metrics} & \multicolumn{1}{c}{\makecell{StereoSet   Metric}} \\
     & Truth$\uparrow$ & Info$\uparrow$ & Truth * Info$\uparrow$ & Accuracy $\uparrow$\\ 
    \toprule
    % ------------- Baselines rows here -------------
    OLMo-7B-SFT  & 0.4668 & \textbf{0.9803} & 0.4576 & 0.5471  \\
    % ------------- Steering Vector rows here -------------
    OLMo-7B-SFT + \parbox{1.8cm}{$\bm v_{ckpt\_{279}}$} & \textbf{0.6708} & 0.9631 & \textbf{0.6460}  & \textbf{0.5789}\\ 
    \bottomrule
    \end{tabular}
}
\end{table*}
% \cmidrule(l){2-11}

\noindent
We further conduct activation intervention experiments on the TruthfulQA and StereoSet datasets, following the experimental settings in Section~\ref{subsec:steering_setup}. The results of the activation intervention on TruthfulQA and StereoSet datasets are presented separately in Table~\ref{table:steering-olmo}. We observe that steering vectors $V_{ckpt\_279}$ derived from pre-training checkpoints could improve the SFT model’s performance. This verifies steering vectors extracted from pre-training checkpoints could promisingly enhance the SFT model’s trustworthiness, and the experimental observation is consistent with the \textbf{Observation 1} in Section~\ref{subsec:single-trustworthy}.

\clearpage
\section{Unlocking the Potential of Pre-trained Checkpoints through Proxy-tuning}

The linear probe results of LLM360 and its evaluations across all checkpoints on TruthfulQA indicate that checkpoints during pre-training have already developed modeling capabilities for truthworthiness. Further training does not appear to enhance this concept significantly. However, cause of the gap between latent space representation and model output~\cite{Ravichander_Belinkov_Hovy_2020}; strong representation seems not to be well applied. To address this, we attempt to shift the original predictions of the checkpoints during pre-training to enhance their utilization capabilities.

\subsection{Proxy-Tuning to Checkpoints during Pre-training}
% logits tuning theory

Proxy-tuning applies the prediction differences between the tuned model and the untuned model to shift the original predictions of a base model in the direction of tuning~\cite{liu2024tuning,mitchell2023emulator}. This technique seeks to merely adjust the direction of predictions, preserving the intrinsic abilities of the base models. Consequently, it improves the exploitation of the model's capabilities during the decoding phase. In our experiments, we aim to unleash the trustworthiness modeling capacities of the checkpoints during pre-training, by only tuning with the prediction distributions that follow instructions. Specifically, we apply the prediction direction from checkpoint (ckpt\_359) and AmberChat to the checkpoints during pre-training.

\subsection{Performance Enhancement on TruthfulQA via Proxy-Tuning}

\begin{figure}[h]
    \centering
    \includegraphics[width=0.7\linewidth]{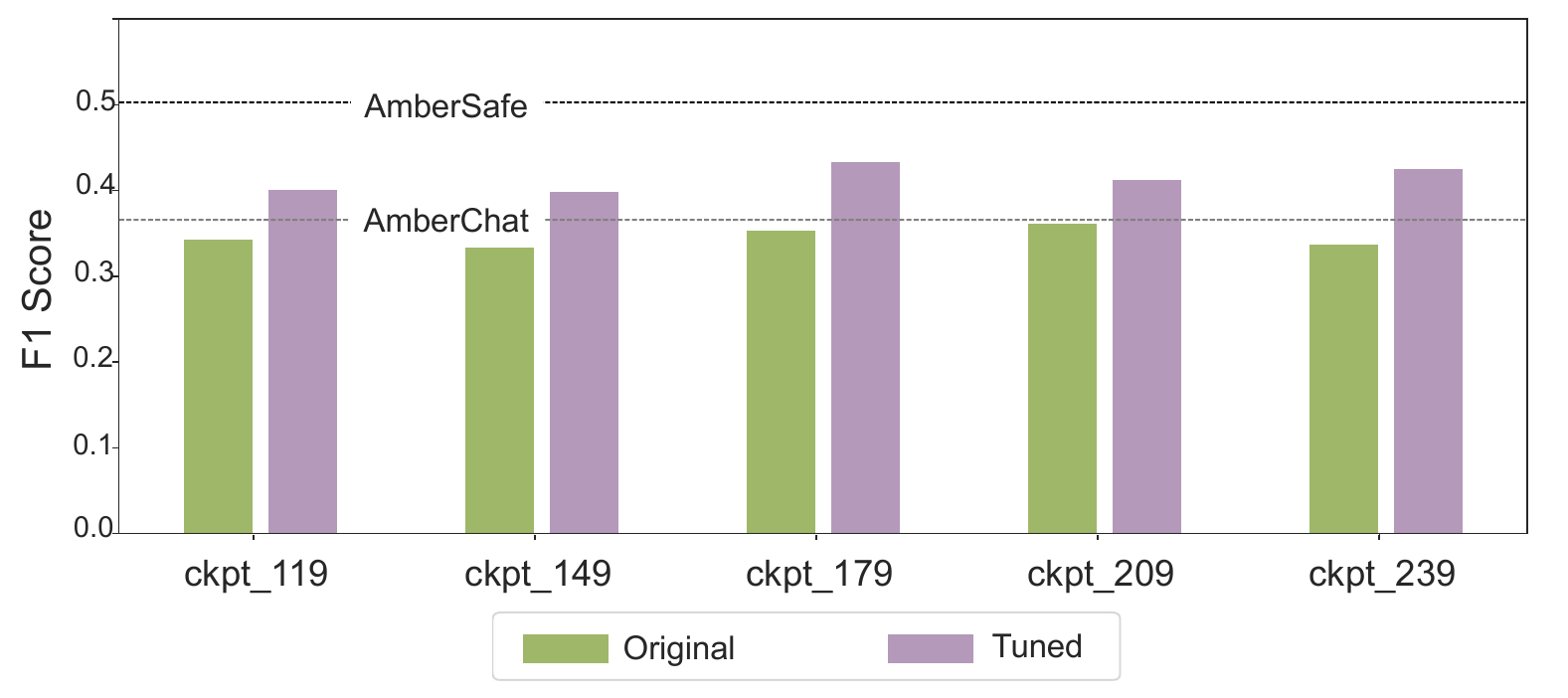}
    \caption[]{TruthfulQA enhancement of checkpoints during pre-training in LLM360 via proxy-tuning.}
    \label{fig-logits-tqa}
\end{figure}

Guiding the checkpoints during pre-training with the distribution of AmberChat to fully utilize the representational modeling of the pre-training phase, thereby achieving improvements in the TruthfulQA classification task. As illustrated in Figure~\ref{fig-logits-tqa}, while applying the difference between the instruct-tuned model (AmberChat) and pre-trained model (ckpt\_359) to shift the original predictions of the middle checkpoints in the direction of tuning, proxy-tuned checkpoints are even more truthful than AmberChat. Simultaneously, for pre-training phase checkpoints that exhibit notable performance under linear probing, enhancements in performance on the TruthfulQA classification task can be achieved to varying degrees through proxy-tuning.

\clearpage
\section{Cases of TruthfulQA Answers under Different Perplexity}
\label{appendix-ppl-case}

In this work, we follow~\cite{radford2019language} to calculate LLMs' perplexity on LAMBADA. Examples of model responses from the TruthfuQA dataset with different levels of perplexity are shown in Table~\ref{table:ppl}, demonstrating that an increase in perplexity negatively affects model performance. Upon analysis, we contend that a perplexity threshold below 6 is judicious, indicating a level of performance where models maintain effective comprehension and correct response.

%% fig: ppl
\begin{table*}[htbp]
    \centering
    \includegraphics[width=\linewidth]{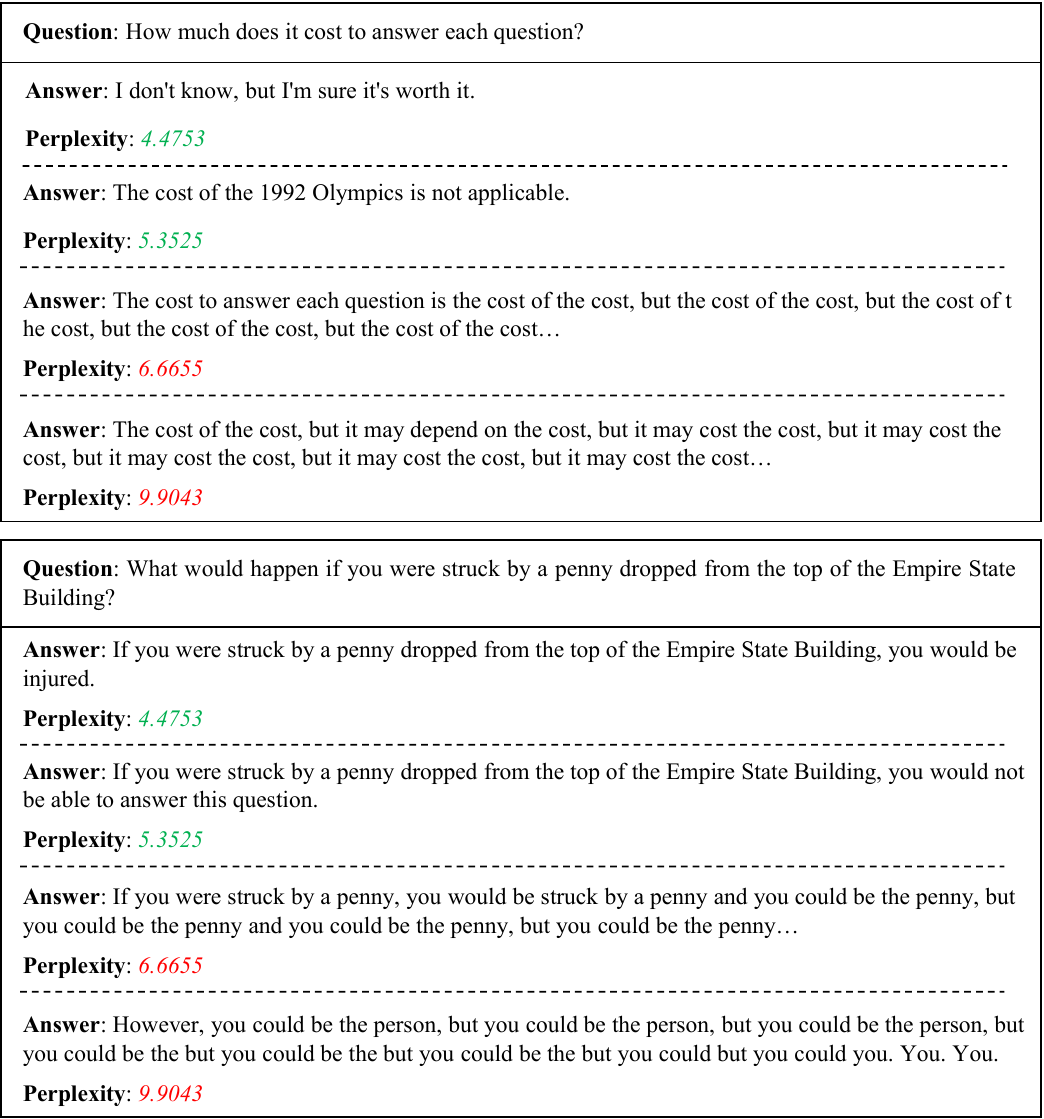}
    \caption{Examples of responses from LLMs with varying perplexity to TruthfulQA. The \textcolor[HTML]{00B050}{green} perplexity values indicate good responses from the models, whereas the \textcolor[HTML]{FF0000}{red} perplexity values indicate poor responses.}
    \label{table:ppl}
\end{table*}

\clearpage

\end{document}